\def\year{2022}\relax
\documentclass[letterpaper]{article} 
\usepackage{aaai22}  
\usepackage{times}  
\usepackage{helvet}  
\usepackage{courier}  
\usepackage[hyphens]{url}  
\usepackage{graphicx} 
\usepackage{natbib}  
\usepackage{caption} 
\DeclareCaptionStyle{ruled}{labelfont=normalfont,labelsep=colon,strut=off} 
\usepackage{algorithm}
\usepackage{algorithmic}

\usepackage{newfloat}
\usepackage{listings}
\floatstyle{ruled}
\newfloat{listing}{tb}{lst}{}
\floatname{listing}{Listing}

\usepackage{soul}
\usepackage{epstopdf}
\usepackage[utf8]{inputenc}
\usepackage{url,textcomp}  
\usepackage{tablefootnote}
\usepackage{verbatim}
\usepackage{mdwlist}
\usepackage{tabularx}
\usepackage{amsmath,amsfonts,amsthm,bm,amssymb}
\usepackage{graphicx}
\usepackage{enumitem}
\usepackage{array}
\usepackage{arydshln}
\usepackage{mathtools}

\usepackage{xspace}
\usepackage{xcolor}
\usepackage{graphicx}
\usepackage{subcaption}
\usepackage{amsfonts}
\usepackage{algorithm}
\usepackage{booktabs}

\usepackage{stmaryrd}
\newcommand*\concat{\mathbin{\|}}
\def\taggen{\textsc{TagGen}\xspace}
\def\dymond{\textsc{Dymond}\xspace}
\def\namemodel{\textsc{Tigger}\xspace}
\def\sage{\textsc{GraphSAGE}\xspace}

\usepackage{todonotes}
\usepackage[toc,page]{appendix}
\usepackage{diagbox}
\usepackage{multirow}
\usepackage{array}
\usepackage{makecell}
\usepackage{fmtcount} 
\usepackage{array,multirow}
\usepackage{tablefootnote}

 \setlist{nolistsep,leftmargin=*}

\newcommand{\ch}{\mathbf{h}\xspace}
\newcommand{\co}{\mathbf{o}\xspace}
\newcommand{\cv}{\mathbf{v}\xspace}
\newcommand{\cf}{\mathbf{f}\xspace}

\newcommand{\CG}{\mathcal{G}\xspace}
\newcommand{\CV}{\mathcal{V}\xspace}
\newcommand{\CE}{\mathcal{E}\xspace}

 \newtheorem{defn}{\textbf{Definition}}
  \newtheorem{prob}{\textbf{Problem}}
  \newtheorem{ex}{\textbf{Example}}
  \newtheorem{thm}{\textbf{Theorem}}

\title{TIGGER: Scalable Generative Modelling for Temporal Interaction Graphs}

\author {
    Shubham Gupta,
    Sahil Manchanda,
    Srikanta Bedathur,
    Sayan Ranu
}
\affiliations {
    Department of Computer Science and Engineering \\ 
    Indian Institute of Technology, Delhi\\
    \{shubham.gupta,sahil.manchanda,srikanta,sayanranu\}@cse.iitd.ac.in
}

\begin{document}

\maketitle
\vspace{-0.20in}
\begin{abstract}

\vspace{-0.10in}
There has been a recent surge in \textit{learning} generative models for graphs. While impressive progress has been made on static graphs, work on generative modeling of \textit{temporal} graphs is at a nascent stage with significant scope for improvement. First, existing generative models do not scale with either the time horizon or the number of nodes. Second, existing techniques are \textit{transductive} in nature and thus do not facilitate knowledge transfer. Finally, due to their reliance on one-to-one node mapping from source to the generated graph, existing models leak node identity information and do not allow \textit{up-scaling/down-scaling} the source graph size. In this paper, we bridge these gaps with a novel generative model called \namemodel. \namemodel derives its power through a combination of \textit{temporal point processes} with \textit{ auto-regressive modeling} enabling both transductive and inductive variants. Through extensive experiments on real datasets, we establish \namemodel generates graphs of superior fidelity, while also being up to $3$ orders of magnitude faster than the state-of-the-art. 
\end{abstract}

\vspace{-0.10in}
\section{Introduction and Related Work}
\label{sec:intro}
Modelling and generating graphs find applications in various domains such as drug discovery~\cite{popova2019molecularrnn,li2018multi}, anomaly detection~\cite{ranu2009graphsig}, data augmentation~\cite{pmlr-v80-bojchevski18a}, and data privacy~\cite{dataprivacy}. 
Initial works on graph generative modelling relied on making prior assumptions about the graph structure. Examples include \textit{Erdős-Rényi}~\cite{Karonski1997} graphs, \textit{small-world} models~\cite{small_world}, and \textit{scale-free} graphs~\cite{albert2002statistical}.  Recently, learning-based algorithms have been developed that circumvent this limitation~\cite{you2018graphrnn, goyal2020graphgen,popova2019molecularrnn,cao2018molgan,liao2019gran}. Specifically, these algorithms directly learn the underlying hidden distribution of  graph structures from training data.

Unfortunately, most of the learning-based generative models are limited to static graphs. In today’s world, there is an abundance of graphs that are \textit{temporal} in nature. Examples include financial transactions~\cite{kumar2016edge,8038008}, online shopping~\cite{amazon_dataset}, community interaction graphs like Reddit~\cite{Liu-2019-sampling}, and user behaviour networks~\cite{foursquare}. The interactions (edges) between nodes in a temporal graph are timestamped and the structure of these graphs change with time. The key challenge in generative modelling is therefore to learn the rules that govern their evolution over the time horizon~\cite{Michail2015}. 

\taggen~\cite{TagGen} models temporal graphs by converting them into equivalent static graphs by     combining node-ids with each of their interaction edge timestamps, and connecting only those  nodes in the resulting static graph that satisfy a specified temporal neighbourhood constraint. They perform random walks on this transformed graph, which are then modified using heuristic local operations to generate many synthetic random walks. Finally, the synthetic random walks that are classified by a discriminator as real random walks are collected and combined to construct the generated temporal graph.
More recently, \dymond~\cite{dymond} presented a non-neural, 3-node motif based approach for the same problem. They assume that each type of motif follows a time-independent exponentially distributed arrival rate and learn the parameters to fit the observed arrival rate. 

These approaches suffer from the following limitations:

    \noindent
    $\bullet$ {\textbf{Weak Temporal Modelling:} \dymond makes two key assumptions: first, the arrival rate of motifs is exponential; and second, the structural configuration of a motif remains the same throughout the time horizon being modeled on. Both these assumptions do not hold in practice -- motifs themselves may evolve with time and could arrive with time-dependent rates. This leads to poor fidelity of structural and temporal properties of the generated graph. \taggen, on the other hand, does not model the graph evolution rate explicitly. It assumes that the timestamps in the input graph are discrete random variables prohibiting \taggen from generating new(unseen in source graph) timestamps. 
More critically, the generated graph duplicates a large portion of edges from the source graph -- our experiments found upto $80\%$ edge overlap between the generated and the source graph. While the design choices of \taggen generate graphs that exhibit high fidelity of graph structural and temporal interaction properties, unfortunately it achieves them by generating graphs that are largely indistinguishable from the source graph due to their poor modelling of interaction times.} 
    
    \noindent
    $\bullet$ \textbf{Poor Scalability to Large Graphs:}  Both \taggen and \dymond are limited to graphs where the number of nodes are less than $\approx$10000  and the number of unique timestamps are below $\approx$200. However, real graphs are not only of much larger size, but also grow with significantly high interaction frequency \cite{Paranjape_2017}. In such scenarios, the key design choice of \taggen to convert the temporal graph into a static graph, 
fails to scale to long time horizons 
 since the number of nodes in the resulting static graph multiplies linearly with the number of timestamps. Further, \taggen also requires the computation of the inverse of an $N' \times N'$ matrix, where $N'$ 
is the number of nodes in the equivalent static graph to impute node-node similarity. This leads to the quadratic increase in memory consumption and even higher cost of matrix inversion, thus making \taggen not scalable.
 On the other hand, \dymond has an $O(N^3T)$ complexity, where $N$ is the number of nodes and $T$ is the number of timestamps. In contrast, the complexity of the algorithm we propose is in $O(NM)$ for a graph with $N$ nodes and $M$ timestamped edges, and is independent of the time horizon length.

    \noindent
    $\bullet$ \textbf{Lack of Inductive Modelling:} Inductivity allows transfer of knowledge to unseen graphs~\cite{hamilton2018inductive}. In the context of graph generative modelling, inductive modelling is required to \textbf{(1)} upscale or downscale the source graph to a generated graph of a different size, and \textbf{(2)} prevent leakage of node-identity from the source graph. Both \taggen and \dymond rely on one-to-one mapping from source graph node ids to the generated graph and hence are non-inductive. 





\textbf{Contributions:} The proposed generative model, \namemodel (\underline{\textbf{T}}emporal \underline{\textbf{I}}nteraction \underline{\textbf{G}}raph \underline{\textbf{GE}}ne\underline{\textbf{R}}ator), addresses the above mentioned gaps in existing literature through the following novel contributions:
\begin{itemize}

\item \textbf{Assumption-free Modelling:} We utilize \textit{intensity-free} \textit{temporal point processes} (TPP)  TPPs~\cite{shchur2020intensityfree} to jointly model the underlying distribution of node interactions and their timestamps through \textit{temporal random walks}.  Our modelling of time is \textit{assumption-free} as we fit a \textit{continuous} distribution over time. This allows \namemodel to generate timestamps that were not even present in the input graph. Moreover, this empowers \namemodel to sample interaction graphs for future-timestamps. Thus, \namemodel is capable of up-sampling/down-sampling  in the temporal dimension.
 
\item \textbf{Inductive Modelling:} 
\namemodel supports inductive modelling through a novel \textit{multi-mode decoder} that learns the distribution over node embeddings instead of learning distribution over node IDs. In addition,  through the usage of a \textsc{Wgan} \cite{wgan}, we support up-sampling/down-sampling of generated graph size. Thus, in contrast to \dymond and \taggen, \namemodel is capable of generating graphs of arbitrary sizes without leaking information from the source graph -- potentially useful in many privacy-sensitive applications. 
\item \textbf{Large-scale Empirical Evaluation:} Extensive evaluation over five large, real temporal graphs with up to to millions of timestamps comprehensively establishes that \namemodel breaks new ground in terms of its scalability, while also ensuring superior fidelity of structural and temporal properties of the generated graph.

 \end{itemize}


\begin{figure*}[t]
    \centering
    \vspace{-0.20in}
        \includegraphics[width=6.5in]{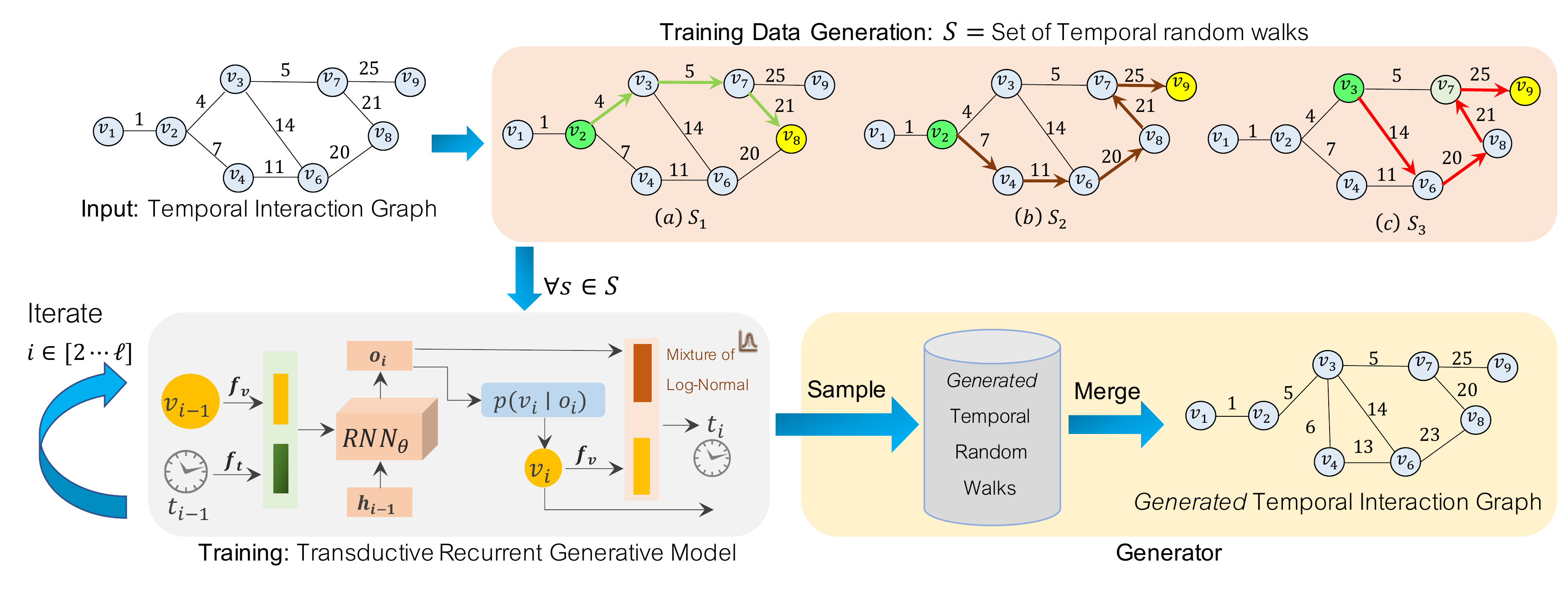}
        \vspace{-0.15in}
        \caption{{Pipeline of \namemodel in transductive modelling. The highlighted nodes in the Training Data Generation component indicates the start (green) and the ending nodes (yellow) of random walks for different length $\ell$.}}
        \label{fig:tigger}
        \vspace{-0.20in}
\end{figure*} 
\vspace{-0.10in}
\section{Problem Formulation}

\begin{defn}[Temporal Interaction Graph]
\label{def:def_tempInt}
A temporal interaction graph is defined as $\mathcal{G} = (\mathcal{V},\mathcal{E})$ where $\mathcal{V}$ is a set of $N$ nodes and $\mathcal{E}$ is a set of $M$ temporal edges $\{(u,v,t) \mid u,v  \in \mathcal{V},t \in [0, T] \}$. $T$ is the maximum time of interaction. 
\end{defn}
\noindent %

\begin{prob}[Temporal Interaction Graph Generator]\hfill\newline
\noindent
\textbf{Input:} A temporal interaction graph $\mathcal{G}$. \\
\textbf{Output:} Let there be a hidden joint distribution of structural and temporal properties from which given $\mathcal{G}$ has been sampled. Our goal is to learn this hidden distribution. Towards that end, we want to learn a generative model $p(\mathcal{G})$ that maximizes the likelihood of generating $\mathcal{G}$. This generative model, in turn, can be used to generate new graphs that come from the same distribution as $\mathcal{G}$, but not $\mathcal{G}$ itself.
\end{prob}

The above problem formulation is motivated by the \textit{one-shot generative modelling} paradigm i.e., it only requires one temporal graph $\mathcal{G}$ to learn the hidden \textit{joint} distribution of structural and temporal interaction graph properties. Defining the joint distribution of temporal and structural properties is hard. In general, these properties are characterized by inter-interaction time distribution and evolution of static graph properties like degree distribution, power law exponent, no. of connected components, largest connected component, distribution of pair wise shortest distances, closeness centrality etc. Typically, a generative model optimizes over one of these properties under the assumption that the remaining properties are correlated and hence would be implicitly modeled. For example, \dymond uses small structural motifs and \taggen uses random walks over the transformed static graph. In our work, we perform \textit{temporal random walks}, which are then modeled using \textit{point processes}.

\vspace{-0.10in}
\section{\namemodel}
Fig.~\ref{fig:tigger} presents the pipeline of \namemodel. Given a source graph, we decompose it through temporal random walks. These random walks are modeled using a recurrent generative neural model. Once the model is trained, it is used to generate synthetic temporal random walks, which are finally merged to form the generated temporal interaction graph. We next formalize each of these sub-steps.

\vspace{-0.10in}
\subsection{Training Data: Temporal Random Walks}
\begin{defn}[Temporal Neighborhood] 
The temporal neighbourhood of a node $v$ at time $t$ contains all edges with a higher time stamp. Formally,
\vspace{-0.05in}
\begin{equation}
\label{eq:temp_nbrhood}
\nonumber
\mathcal{N}_{t}(v) = \left\{e \;\mid\; e=(v, u, t') \in \mathcal{E} \wedge t < t' \right\}
\end{equation}
\end{defn}
\begin{defn}[Temporal Random Walk]
\label{def:temprandom_walk}Given a node $v$ and time $t$, an $\ell$-length temporal random walk starts from $v$ and takes $\ell$ jumps through an edge in the temporal neighborhood of the current node. More formally, it is a sequence of tuples $S =\{s_1,\cdots,s_{\ell}\}$, where each tuple $s\in S$ is a $(node,time)$ pair such that, $s_1=(v,t)$ and $\forall i\in [2,\ell]$ the edge $\left(s_{i-1}.v, s_{i}.v, s_{i}.t\right) \in \mathcal{N}_{s_{i-1}.t}(s_{i-1}.v) $. A walk ends after taking $\ell$ jumps or if $\mathcal{N}_{s_{i-1}.t}(s_{i-1}.v)=\emptyset$.
\end{defn}

 Since each jump is constrained to edges within the temporal neighborhood, it is guaranteed that $s_i.t>s_{i-1}.t$. To capture the temporal characteristics, the probability of jumping through edge $e\in \mathcal{N}_{s_{i}.t}(s_{i}.v)$ decreases exponentially with time gap from $(s_{i}.t)$. More formally,

\vspace{-0.10in} 
{\footnotesize
\begin{alignat}{1}
\nonumber
    p(e =(s_i.v,u,t) \mid s_i) =
    &\frac{\exp\left(s_{i}.t-t\right) }{\sum \limits_{e'=(s_i.v,u',t') \in \mathcal{N}_{s_{i}.t}(s_{i}.v)} \exp\left( s_{i}.t-t'\right)}
\end{alignat}
}
Note that $s_i.t<t$ and hence a smaller gap leads to increased chances of being sampled. While we exponentiate the time gap, other functions, such as linear, may also be used. We use exponentiation due to superior empirical results. A random walk starts from an edge chosen uniformly at random.  Examples of temporal random walks are shown in the Data Generation component of Fig.~\ref{fig:tigger}.
%
 In Table \ref{tab:notation} in appendix, we summarise all the notations used in our work. As per convention, we use boldface symbols to denote learnable vectors and weight matrices.
\vspace{-0.05in}
\subsection{Modelling Temporal Random Walks}
We train a generative model $p(\mathcal{S})$ on a set $\mathcal{S}$ of temporal random walks. Formally, 
\vspace{-0.05in}
\begin{alignat}{2}
\nonumber
p(\mathcal{S}) &= \prod_{S\in\mathcal{S}} p(S)\\
\text{where, } p(S) &= p(s_1, \ldots , s_{\ell}) 
\label{eq:probs}
\end{alignat}
Owing to the \textit{auto-regressive} nature of a sequence, we express $p(S)$ as the product of the conditionals.

\vspace{-0.20in}
\begin{alignat}{1}
    \nonumber
    p(S) &= p(s_1)\prod_{i=2}^{{\ell}} p(s_{i} \mid (s_1,\ldots, s_{i-1}))
\end{alignat}
We simplify this conditional by decomposing as follows. 
\vspace{-0.05in}
\begin{alignat}{2}
p(S) &= p(s_1)\prod_{i=2}^{{\ell}}  p(s_{i}.v \; | \; (s_1,\ldots , s_{i-1}))\\
   & \times p(s_{i}.t \mid (s_{i}.v,(s_1, \ldots , s_{i-1})))
\label{eq:decompose_prob}
    \end{alignat}

To learn the above conditional distribution, we utilize a \textit{recurrent neural network (RNN)} based generator. Formally,
\vspace{-0.05in}
\begin{alignat}{2}
\nonumber
\ch_i &{=}rnn^{hidden}_{\theta}(\ch_1,(s_1,\ldots, s_{i-1})){=}rnn^{hidden}_{\theta}(\ch_{i-1},s_{i-1})\\
\nonumber
\co_i &{=} rnn^{output}_{\theta}(\ch_1,(s_1,\ldots, s_{i-1})){=}rnn^{output}_{\theta}(\ch_{i-1},s_{i-1})
\end{alignat}
Here, $rnn^{output}_{\theta}(\ch_{i-1},x)$ is the output of RNN cell and $rnn^{hidden}_{\theta}(\ch_{i-1},x)$ is the updated hidden state. Both $\ch_i$ and $\co_i$ are vectors and we initialize $\ch_1 = \textbf{0}$. Semantically,  $\co_i$ captures the \textit{prior} to predict the next node $v_i$ in the temporal random walk. More formally,   $p(S)$ in Eq.~\ref{eq:decompose_prob} is re-written as:
\vspace{-0.07in}
\begin{equation}
p(S) = p(s_1)\prod_{i=2}^{\ell} p(s_{i}.v \mid \co_{i})*p(s_{i}.t \mid s_{i}.v,\co_{i})
\label{eq:main_eq}
\end{equation}

In the following sections, we discuss the internals of the RNN, and formulate how exactly Eq.~\ref{eq:main_eq} is learned. We develop two procedures: first is a transductive learning algorithm, and the second is an inductive model.%

\noindent
\textbf{Transductive Recurrent Generative Model:}
Given a sequence of node and time pairs ${s_1 \ldots , s_\ell}$, we transform  $s_i.v$ and $s_i.t$ to vector representations $\cf_v(s_i.v) \in \mathcal{R}^{d_V} ,\ \cf_t(s_i.t) \in \mathcal{R}^{d_T}$ respectively. 

First, we transform the node ids to a vector using $\cf_v(v) = \textbf{W}_v\cv$ where $\textbf{W}_v \in \mathcal{R}^{d_V}*\mathcal{R}^{N}$ is a learnable weight matrix and $\cv\in\mathcal{R}^{1\times N}$ is \textit{one-hot encoding} of the node ID of $v$. Next, to learn vector representation of time $t \in \mathcal{R}$, we use following  \textsc{Time2Vec}~\cite{kazemi2019time2vec} transformation.
\vspace{-0.05in}
\begin{equation}
\cf_t(t)[r]= 
\begin{cases}
    \omega_r\cdot t + \zeta_r,& \text{if } r=0 \\
    \sin{(\omega_r \cdot t + \zeta_r)}, & 1 \leq r < d_T
\end{cases}
\label{eq:time2vec}
\end{equation}
where $\omega_1,\omega_2,\ldots, \omega_{d_T}, \zeta_1,\zeta_2,\ldots, \zeta_{d_T} \in \mathcal{R}$ are trainable weights and shared across each pair of the input sequence. $r$ is the index of $\cf_t(t)$.
After embedding both $s_{i-1}.v$ and $s_{i-1}.t$, we concatenate them resulting in a vector of $\mathcal{R}^{d_V+d_T}$ dimension. This vector is fed into the RNN cell along with  $\ch_{i-1}$ which outputs $\co_{i}$  and $\ch_{i}$. We represent $p(s_{i}.v\mid \co_{i})$ in Eq.~\ref{eq:main_eq} as multinomial distribution over $v \in \mathcal{V}$ parameterized by $\theta_{v}$.
\begin{align}
\nonumber & \hspace{-0.1in}p(s_i.v =v \mid \co_{i}) = \theta_{v}(\co_i) \\
\nonumber
& = \theta_{v}(rnn_{\theta}^{output}(\ch_{i-1},(s_{i-1}.v,s_{i-1}.t))) \\
\nonumber
& = \theta_{v}(rnn_{\theta}^{output}(\ch_{i-1},(\cf_v(s_{i-1}.v)\concat\cf_t(s_{i-1}.t)))) \\ 
& =\frac{exp(\textbf{W}^O_v\co_i)}{\sum_{\forall u\in\CV} exp(\textbf{W}^O_u\co_i)}
\label{eq:nodeprob}
\end{align}
where $\textbf{W}^O_v \in \mathcal{R}^{1*d_O},\: \forall v \in \mathcal{V}$ is a node-specific learnable weight vector. $d_O$ is the dimension of $o_i$.

\par
\textit{Temporal point processes (TPP)} are de-facto models for modelling distributions of continuous, inter-event time over discrete events in event sequences $\left\{(e_0,t_0),(e_1,t_1) \ldots (e_n,t_n)\right\}$. TPPs are generally defined using \textit{conditional intensity function} $\lambda(t)$.
$$\lambda(t) = \frac{p(t \mid \textbf{H}_{t_n})}{1-F(t \mid \textbf{H}_{t_n})}$$ 
Here, $p(t\mid \textbf{H}_{t_n})$ is the probability distribution of next event time $t$ after observing events till time $t_n$. $F(t)$ is the cumulative probability distribution corresponding to $p$. $\textbf{H}_{t_n}$ is the summary of events till time $t_n$. $\lambda(t) $ is the expected number of events around infinitesimal interval $[t,t+dt]$ given the history before $t$. It results in following probability distribution $p$ for next event time \cite{rizoiu2017tutorial}.
\begin{equation}
    \nonumber
p(t \mid \textbf{H}_{t_n}) = \lambda(t)\exp(- \int_{t_n}^{\infty} \lambda(x) \,dx )
\end{equation}
\vspace{-0.01in}
Resulting log likelihood contains integral due to $p(t)$ which needs to be estimated using \textit{Monte-Carlo sampling} \cite{mei2017neural} leading to high variance, unstable updates during training and high computation cost \cite{omi2020fully}. Motivated by strong performance on event time prediction task by \cite{shchur2020intensityfree}, we adopt their TPP formulation, which directly defines $p(t)$ as mixture of \textit{log normal} distribution instead of deriving it from $\lambda(t)$. From Eq.~\ref{eq:main_eq}, 
\vspace{-0.05in}
\begin{equation}
    \begin{gathered}
    p(s_{i}.t \mid s_{i}.v,\co_i) = p(s_{i}.t - s_{i-1}.t \mid s_{i}.v,\co_i) \\\quad\quad\quad\;= \theta_t(\Delta t \mid s_{i}.v,\co_i)\\ = \sum_{c=1}^{C} \phi_c^C \frac{1}{\Delta t \sigma_{c}^C \sqrt{2 \pi}} \exp (-\frac{(\log \Delta t-\mu_{c}^{C})^{2}}{2 (\sigma_{c}^{C})^2})
    \end{gathered}
    \label{eq:timeprob}
\end{equation}
\noindent
where $\Delta t$ is time difference between $s_i$ and $s_{i-1}$, $p(t)$ is parameterized by $\theta_t$ and $\mu_c^{C},\sigma_c^{C}, \phi_c^{C}$ are parameters of $\theta_t$. 
\begin{equation*}
\begin{gathered}
\mu_c^{C} {=} \textbf{W}^{\mu C}_{c}(f_v(s_{i}.v)\concat \co_i),\;  \sigma_c^{C} {=} exp(\textbf{W}^{\sigma C}_{c}(f_v(s_{i}.v)\concat \co_i)) \\
\quad  \phi_c^{C} =\frac{exp(\textbf{W}^{\phi C}_c(f_v(s_{i}.v)\concat \co_i))}{\sum_{j=1}^{C} exp(\textbf{W}^{\phi   C}_j(f_v(s_{i}.v)\concat \co_i))}
\end{gathered}
\end{equation*}
Moreover, $C$ is no. of components in the \textit{log normal} mixture distribution and 
$\textbf{W}^{{\mu C}}_c,\textbf{W}^{\sigma C}_c, \textbf{W}^{\phi C}_c \in \mathcal{R}^{(d_V+d_O)}, \; \forall c \in \{1..C\}$. Note that every components' learnable weights are shared across each time-stamp in the sequence.

\noindent
\textbf{Training loss:} The loss over the set $\mathcal{S}$ of temporal random walks is derived from Eqs.~\ref{eq:probs}, \ref{eq:main_eq}, \ref{eq:nodeprob} and \ref{eq:timeprob}. Specifically,
\vspace{-0.05in}
\begin{equation}
\begin{aligned}
\nonumber
\mathcal{L} &= -\log(p(\mathcal{S})) = -\sum_{S\in\mathcal{S}} \log(p(S)) \\ &= -\sum_{S\in\mathcal{S}} \log p(s_1) \sum_{i=2}^{\ell} (\log(p(s_i.v \mid \co_i) \\&\quad\quad\quad\quad\quad\quad+ \log(p(s_i.t\mid s_i.v,\co_i)))
\end{aligned}
\end{equation}
In the above loss function, $p(s_1)=1/|\CE|$ since the first edge is chosen uniformly at random. $p(s_i.v \mid \co_i)$ and $p(s_i.t \mid s_i.v,\co_i)$ are computed using Eq.~\ref{eq:nodeprob} and Eq.~\ref{eq:timeprob} respectively.
 A pictorial summary of the training process is available in the training component of Fig.~\ref{fig:tigger}.
\subsubsection{Inductive Recurrent Generative Model:}
The primary distinction between transductive and inductive generative models are the construction of node representation and the procedure of learning next node distribution given the past information in the sequence. In the transductive model, a node is represented by its ID $\in \{1, \ldots, N \}$ in the form of a one-hot vector. In the inductive model, we use a \textit{Graph Convolution Network (GCN)} to embed nodes. 

\par
\textbf{Node Representations:} We first transform the input temporal graph $\mathcal{G}=(\CV,\CE)$ to a static graph $\mathcal{G}^{static}=(\CV,\CE^{static})$, where $\CE^{static}=\{(u,v)\mid \exists (u,v,t)\in \CE\}$. On $\CG^{static}$, we utilize \sage \cite{hamilton2018inductive} to learn \textit{unsupervised} structural node representations. Details can be found in the appendix.

We denote embedding of node $v$ as $\textbf{v} \in \mathcal{R}^{d_{\textbf{V}}}$. Given a temporal walk sequence $S=(s_1, \ldots s_\ell)$, we replace $s_i.v$ with $s_i.\textbf{v} \; \forall i \in \{1,\ldots,\ell\}$. Similar to the transductive variant, in order to learn $p(s_i\mid s_1,\ldots, s_{i-1})$, each $s_{i-1}.\textbf{v}$ and $s_{i-1}.t$ is transformed using $\cf_\textbf{v}(s_{i-1}.\textbf{v}) = \textbf{W}s_{i-1}.\textbf{v}$ where $\textbf{W} \in \mathcal{R}^{d_\textbf{V}} \times \mathcal{R}^{d_\textbf{V}}$ and $\cf_t$ using Eq.~\ref{eq:time2vec}.
Both $\cf_\textbf{v}(s_{i-1}.\textbf{v})$ and $\cf_t(s_{i-1}.t)$ are concatenated, 
which is fed into the RNN cell along with the previous hidden state $\ch_{i-1}$. The RNN outputs $\co_i \in \mathcal{R}^{d_O}$  and $\ch_i \in \mathcal{R}^{d_H}$. These steps are the same as in the transductive variant. 

\begin{figure}[t]
    \centering
    \vspace{-0.10in}
    \includegraphics[width=3.4in]{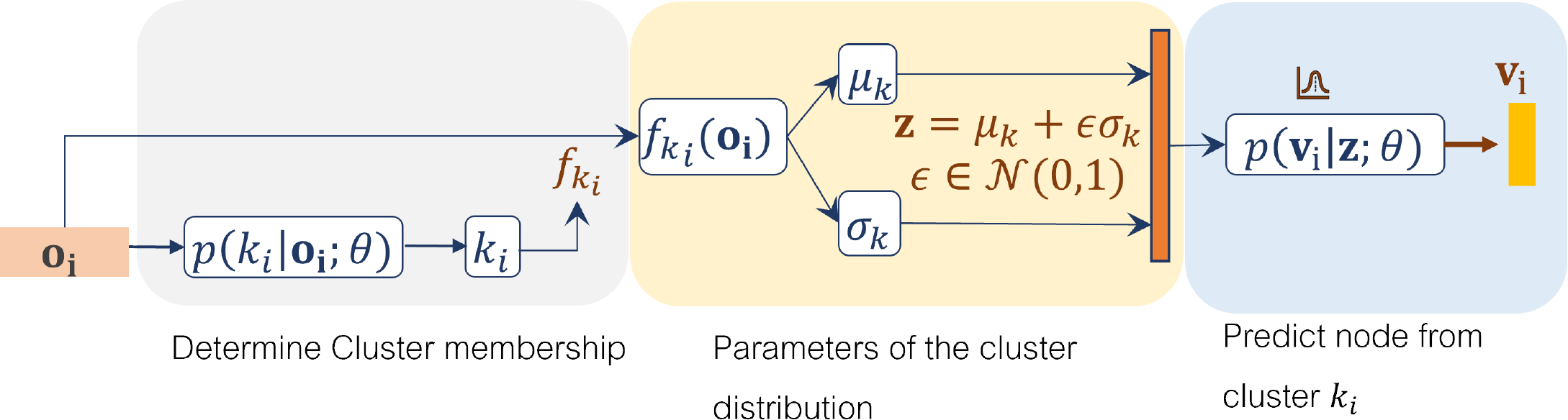}
    \vspace{-0.25in}
    \caption{{Multi-mode decoder}}
    \label{fig:cve}
    \vspace{-0.1in}
\end{figure}

\textbf{Multi-mode node embedding decoder:} Owing to working with node embeddings, the objective of the RNN is to predict the next node embedding instead of a node ID (in addition to the timestamp). Towards that end, we develop a multi-mode node embedding decoder. Fig.~\ref{fig:cve} presents the internals. The decoder has three distinct semantic phases. We explain them below.


We first note that node embeddings of a graph may not  follow a uni-modal distribution since real-world graphs are known to have communities. The presence of communities would create a multi-modal distribution~\cite{hamilton2018inductive}. To model this distribution, we perform $K$-means clustering on the node embeddings; each cluster would correspond to a community. The appropriate value of $K$ may be learned using any of the established mechanisms~\cite{han2011data}.
Next, we design a \textit{multi-mode decoder} that operates in two steps: first, it predicts the cluster that the next node embedding belongs to, and then predicts the node embedding from that cluster. 

Formally, we would like to the learn probability distribution $p(k_i =k\mid \co_i)$, where $k_i$ denotes the cluster membership of next node $s_{i}.v$ and $k\in [1,\cdots,K]$. From this distribution, $k_i$ is sampled. 
Given cluster $k_i$, we next sample a vector $\mathbf{z}$ from  $p(\textbf{z} \mid \co_i,k_i)$. Then, $s_{i}.\textbf{v}$ is sampled from $p(s_{i}.\textbf{v} \mid \textbf{z})$. Since, we need to learn the distribution of $s_{i}.\textbf{v}$ given $\co_i$ and $k_i$, we introduce a latent random variable $\textbf{z}$ in the multi-mode decoder. Mathematically, 
\begin{equation}
\begin{aligned}
    p(s_{i}.\textbf{v} \mid \co_i) &= p(k_i \mid \co_i) \int_{\textbf{z}}p(\textbf{z} \mid \co_i,k_i)p(s_{i}.\textbf{v} \mid \textbf{z}) d\textbf{z} \\ &=  p(k_i \mid \co_i)\mathbb{E}_{ \textbf{z} \sim p(\textbf{z} \mid \co_i,k_i) }[p(s_{i}.\textbf{v} \mid \textbf{z})]
    \label{eq:inductive_prob}
    \end{aligned}
\end{equation}
where,
\vspace{-0.1in}
\begin{equation}
   \begin{gathered}
    \underset{\hspace{-0.25in}\forall k \in \{1, \ldots, K\}}{p(k_i \mid \co_i)} = \theta_{k}(\co_i)= \frac{\exp\left(\textbf{W}^{K}_{k}  \co_i\right)}{\sum_{j=1}^{j=K}\exp\left(\textbf{W}^K_{j}\co_
    i\right)}  \\ p(\textbf{z} \mid \co_i,k_i) = \mathcal{N}\left(\boldsymbol{\mu}^{K}_{k_i},\left(\boldsymbol{\sigma}^{K}_{k_i}\right)^2\right) \\ p(s_i.\textbf{v} \mid \textbf{z}) = \mathcal{N}\left(\boldsymbol{\mu}^{Z},\left(\boldsymbol{\sigma}^{Z})^2\right)\right) \\
    \boldsymbol{\mu}^{K}_{k_i} = \textbf{W}^{\mu K}_{k_i} \co_i \quad \boldsymbol{\sigma}^{K}_{k_i} = \exp\left(\textbf{W}^{\sigma K}_{ k_i}\co_i\right) \\  \boldsymbol{\mu}^{Z} = \textbf{W}^{\mu Z}\textbf{z} \quad \boldsymbol{\sigma}^{Z} = \exp\left(\textbf{W}^{\sigma Z}\textbf{z}\right)
    \end{gathered}
    \label{eq:inductive_prob_1}
\end{equation}
where $\textbf{z} \in \mathcal{R}^{d_{Z}},$ $\boldsymbol{\mu}^K_k,\boldsymbol{\sigma}^K_k,\boldsymbol{\mu}^{Z},\boldsymbol{\sigma}^{Z} \in \mathcal{R}^{d_Z}$,  $\textbf{W}^{\mu Z},\textbf{W}^{\sigma Z} \in \mathcal{R}^{d_Z} \times \mathcal{R}^{d_O}$, $ \textbf{W}^{\mu K}_{k}, \textbf{W}^{\sigma K}_{ k} \in \mathcal{R}^{d_{Z}} \times \mathcal{R}^{d_{Z}}$, $\textbf{W}^{K}_{k} {\in} \mathcal{R}^{1*d_O} \; \forall k \in \{1..K\} $. 
\par
We approximate the $\mathbb{E}$ term in Eq.~\ref{eq:inductive_prob} using the reparameterization trick from the auto-encoding \textit{variational bayes} approach \cite{kingma2014autoencoding} by defining a deterministic function $g$ to represent $\textbf{z} \sim p(\textbf{z} \mid \co_i,k_i)$ as follows:
\vspace{-0.05in}
\begin{equation}
\nonumber
\textbf{z} = g(\boldsymbol{\mu}^{K}_{k_i},\boldsymbol{\sigma}^{K}_{k_i},\varepsilon) = \boldsymbol{\mu}^{K}_{k_i} + \boldsymbol{\varepsilon} \boldsymbol{\sigma}^{K}_{k_i} \quad \boldsymbol{\varepsilon} \sim \mathcal{N}(\textbf{0},\textbf{1}) 
\label{eq:z}
\end{equation}
\begin{equation}
    \begin{gathered}
    \mathbb{E}_{ \textbf{z} \sim p(\textbf{z} \mid \co_i,k_i) }[p(s_{i}.\textbf{v} {\mid} \textbf{z})] {=} \mathbb{E}_{\boldsymbol{\varepsilon} \sim \mathcal{N}(\textbf{0},\textbf{1}) }[p(s_{i}.\textbf{v} {\mid} g(\boldsymbol{\mu}^{K}_{k_i},\boldsymbol{\sigma}^{K}_{k_i},\boldsymbol{\varepsilon}) )]\\ \simeq \frac{1}{L} \sum_{j=1}^{j=L} p(s_{i}.\textbf{v} \mid g(\boldsymbol{\mu}^{K}_{k_i},\boldsymbol{\sigma}^{K}_{k_i},\boldsymbol{\varepsilon}_j)) \quad  \boldsymbol{\varepsilon}_j \sim \mathcal{N}(\textbf{0},\textbf{1})
    \end{gathered}
    \label{eq:z_modified}
\end{equation}
Taking the logarithm of Eq.~\ref{eq:inductive_prob}, substituting the expectation term using Eq.~ \ref{eq:z_modified}, and assuming $L{=}1$\footnote{Motivated by \cite{kingma2014autoencoding}, which shows state-of-the-art empirical results on image generation tasks using L=1}, we get the following: 
\begin{equation}
    \log p(s_{i}.\textbf{v} {\mid} \co_i) {\simeq} \log p(k_i {\mid} \co_i) + \log p(s_{i}.\textbf{v} {\mid} g(\boldsymbol{\mu}^{K}_{k_i},\boldsymbol{\sigma}^{K}_{k_i},\boldsymbol{\varepsilon}))
    \label{eq:inductivenodeprob}
\end{equation}
$p(s_{i}.t {\mid} s_{i}.\textbf{v},\co_i)$ is modelled the same as Eq.~\ref{eq:timeprob} except  $\cf_v(s_{i}.v)$, which is replaced by $\cf_{\textbf{v}}(s_{i}.\textbf{v})$ where $s_{i}.\textbf{v}$ is the vector representation of node $s_{i}.v$.\par
\noindent
\textbf{Training loss} is derived from Eqs.~\ref{eq:probs}, \ref{eq:main_eq}, \ref{eq:timeprob} and \ref{eq:inductivenodeprob}  by substituting the log-probabilities below:
\vspace{-0.05in}
{\small
\begin{equation}
\begin{aligned}
\nonumber
 \mathcal{L} \simeq -\sum_{s \in S}\log p(s_1) \sum_{i=2}^{\ell} (&\log p(s_{i}.\textbf{v} \mid \co_i) + \log p(s_{i}.t\mid s_{i}.\textbf{v},\co_i) \\-&\beta \mathcal{D}_{kl}( p(\textbf{z} \mid \co_i,k_i) \mid\mid \mathcal{N}(\textbf{0},\textbf{1})) ), 
\label{eq:inductive_loss}
\end{aligned}
\end{equation}}
where  $\mathcal{D}_{kl}$ is KL-Divergence. Empirical observations indicate that adding the regularizer on $p(\textbf{z})$ helps in reducing over-fitting. Thus, we have added the KL distance regularization on $p(\textbf{z})$ to restrict its sample space near to the distribution $\mathcal{N}(\textbf{0},\textbf{1})$. Here $\beta \in (0,1)$ is a hyper parameter, which decides the weightage of the regularizer term. $\mathcal{L}$ is then used to learn the model parameters $\textbf{W}, \textbf{W}_k^{\mu K},\textbf{W}^{\sigma K}_k \;\forall k \in \{1\ldots K\},\textbf{W}^{\mu Z},\textbf{W}^{\sigma Z}$ and parameters of $\cf_t,\cf_{\textbf{v}},rnn_{\theta}$.

\subsection{Generating Interaction Graphs}
Once the recurrent generative model is trained over the collection $\mathcal{S}$, we sample synthetic temporal random walks $\mathcal{S}'$ from the trained model. This synthetic collection is then assembled to form the synthetic temporal graph $\mathcal{G}'$. Similar to \cite{TagGen}, from each sequence $S \in \mathcal{S}$, we store the first item $s_1$ and denote $S_1$ as the collection of $s_1$. 

\noindent
\textbf{Transductive model:}
\begin{algorithm}[t]
\caption{Sampling synthetic temporal random walks from a trained transductive recurrent generative model}
\label{alg:strgn}
{\scriptsize
\begin{algorithmic}[1]
	 \REQUIRE $S_1$,  $\cf_v,\cf_t,rnn_\theta,\theta_v \; \forall v\in \mathcal{V},\theta_t$, $\ell'$
	 \ENSURE Synthetic temporal random walks $\mathcal{S}'$
	 \STATE $S^{\prime} = \{\}$
	 \FOR{$s_1 \in S_1$}
	    \STATE $S^{\prime} \gets \{\}$, $(v_1,t_1) \gets s_1$, $\textbf{h}_1 \gets \textbf{0}$ 
	    \FOR{$i \in \{2,3\ldots \ell' \}$}
	        \STATE $\textbf{o}_i,\textbf{h}_i \gets rnn_{\theta}(\textbf{h}_{i-1},(\cf_v(v_{i-1})\concat \cf_t(t_{i-1})))$
	        \STATE $v_i \sim Multinomial(\theta_{v_1}(o_i),\theta_{v_2}(o_i) \ldots \theta_{v_N}(o_i))$ \COMMENT{Sample next node}
	         \STATE $\Delta t \sim \theta_t(t-t_{i-1} \mid v_i,o_i)$ \COMMENT{Sample next time using eq. \ref{eq:time_sample}} \label{lst:line:time_sampling}
	        \STATE $t_i \gets t_{i-1}+\Delta t$
	        \STATE $S^{\prime}=S^{\prime} + (v_{i-1},v_i,t_i)$
	    \ENDFOR 
	 \STATE $\mathcal{S}' = \mathcal{S}'+ S^{\prime}$
	 \ENDFOR
	 \STATE \textbf{Return} $\mathcal{S}'$
\end{algorithmic}}
\end{algorithm}
Alg.~\ref{alg:strgn} explains method to sample synthetic temporal random walks using transductive variant of \namemodel. Specifically, in Alg.~\ref{alg:strgn}, $\Delta t$ in line \ref{lst:line:time_sampling} is sampled using below equation \cite{shchur2020intensityfree}.  
\begin{align}
    \nonumber &\boldsymbol{\phi} \sim  Categorical(\{\phi^C_1\ldots\phi^C_C\}) \\ \Delta t = \exp(&(\boldsymbol{\sigma}^C)^T\boldsymbol{\phi}\varepsilon + (\boldsymbol{\mu}^C)^T\boldsymbol{\phi} ) \quad \varepsilon \sim \mathcal{N}(0,1)  
    \label{eq:time_sample}
    \end{align}
Where $\boldsymbol{\mu}^C=(\mu^C_1\ldots \mu^C_C)$ and $\boldsymbol{\sigma}^C=(\sigma^C_1\ldots \sigma^C_C)$ and $\boldsymbol{\phi}$ is one-hot vector of size $C$.\par
After collecting synthetic temporal random walks $\mathcal{S}'$, we assemble them by maintaining the same edge density as in the original graph within time range $t \in [1,T]$. First, we count the frequency of each temporal occurrence in the synthetic random walks. We denote this as $\alpha (v_i,v_{j},t)$, i.e the frequency of occurrence of node pair $(v_i,v_j)$ at time $t$ in $\mathcal{S}'$. We  denote the set of edges present at time $t$ in  $\mathcal{S}'$  as $\Tilde{E}^{t}$. Now, for each uniquely sampled time stamp $t \in [1,T]$,
we define the distribution of occurrence on node pairs present at time $t$ in synthetic temporal random walks $\mathcal{S}'$ as follows:
\vspace{-0.05in}
\begin{equation}
    p_{v_i,v_j}^t = \frac{\alpha(v_i,v_j,t)}{\sum_{e=(u_i,u_j)  \in \Tilde{E}^t} \alpha(u_i,u_j,t) }
\end{equation}
From this distribution, we keep sampling edges till the edge density of the synthetic graph is same as original graph in the temporal dimension.

\noindent
\textbf{Inductive model:}
Sampling from the inductive version follows a similar pipeline as in the transductive variant; the only difference is the presence of an additional step of mapping the generated node embeddings in the synthetic random walks $S'$ to nodes $v \in \mathcal{G}'$ where $\mathcal{G}'=(\mathcal{V}',\mathcal{E}')$, $|\mathcal{V'}|=N', |\mathcal{E'}|=M'$. $\mathcal{G}'$ is the generated temporal  graph. Note that $N'$ and $M'$ are different from the source graph sizes, and hence allows control over the generated graph size. The pseudocode is provided in Alg. \ref{alg:alg_ind} in appendix. First, we train a \textsc{Wgan}~\cite{wgan} generative model on node embeddings obtained from $\mathcal{G}^{static}$~\cite{ji2021generating}. From the trained \textsc{Wgan} model, we sample $N'$ node embeddings to construct $\CV'$. 
 Finally, we match each embedding in $S'$ to its closest node in $\mathcal{V'}$ using cosine similarity.
 \begin{thm}
  The computation complexities of generating a temporal interaction graph $\CG'=(\CV',\CE')$ through the transductive and inductive versions are $\approx O ( M\times \ell' \times(N  + C))$ and $\approx O ( M\times \ell' \times( K +  C  + N'   ))$  respectively.
 \end{thm} \textsc{Proof.} Provided in Appendix.

\vspace{-0.05in}
\section{Experiments}
\newcolumntype{C}[1]{>{\centering\arraybackslash}m{#1}}
\newcolumntype{P}[1]{>{\centering\arraybackslash\hspace{0pt}}p{#1}}


\begin{table*}[t]
\centering
\vspace{-0.20in}
\resizebox{0.91\textwidth}{!}{

\begin{tabular}{llllp{1.8cm}p{1.2cm}p{1.2cm}p{1cm}p{1cm}p{1cm}p{1cm}p{1.3cm}p{1.2cm}p{0.5cm}p{1cm}p{1cm}p{1cm}}
\toprule
    
    \textbf{Dataset} & $\mathbf{N=|\CV|}$&$\mathbf{M=|\CE|}$&\textbf{T}& \textbf{Method} & \textbf{Time(s)} &  \textbf{\% Edge overlap} & \textbf{Mean degree} &  \textbf{Wedge Count} &  \textbf{Triangle count} & \textbf{PLE} &\textbf{Edge entropy} & \textbf{LCC} & \textbf{NC} & \textbf{Global CF} & \textbf{Mean BC}   & \textbf{Mean CC}  \\ 
    \midrule


    \multirow{4}{1.2cm}{Wiki-Small}  & \multirow{4}{0.9cm}{$1.6K$}&  \multirow{4}{0.9cm}{$2.9K$}&\multirow{4}{0.8cm}{$50$} & \emph{{Median}} & - & - & {{${1.1064}$}}&  ${13.0}$&  ${{0.0}}$ & ${16.4626}$& {{${0.9912}$}}& ${{5.0}}$&{{${44.0}$}}& {{${{0.0}}$}}& {${{0.0}}$}& {{${0.0122}$}} \\
    
    & & & & \dymond & $69120$ & {$\textbf{0.0}$}& {$0.2424$}& {$8.0$}&  {$\textbf{0.0}$}& {$11.426$}&{$0.0075$}& {$2.0$}& {$32.0$}& {$\textbf{0.0}$}& {$0.0005$}& {$0.0264$} \\ 

    & & & & \taggen & $1800$ &        
    {$87.169$}& {$0.0674$}&  {$7.5$}&  {$\textbf{0.0}$}& {$5.6519$}&{$0.0042$}&{$\textbf{1.0}$}& {$\textbf{\textbf{0.0}}$}& {$\textbf{0.0}$}& {$\textbf{0.0}$}&  {$\textbf{0.0004}$} \\

    & & & & \namemodel & \textbf{$\textbf{14}$} & $1.2821$& $\textbf{0.0352}$&  $\textbf{5.0}$&  $\textbf{0.0}$&$\textbf{4.5005}$& $\textbf{0.0039}$&$\textbf{1.0}$& $4.0$& $\textbf{0.0}$& $\textbf{0.0}$& ${0.0014}$\\

\cmidrule(lr){1-17}
    
    \multirow{3}{1.2cm}{ UC Irvine} &\multirow{3}{0.9cm}{$1.8K$}& \multirow{3}{0.9cm}{ $33K$}&\multirow{3}{0.8cm}{$194$} & \textit{Median} & - & - & $1.5714$&  $71.0$ &${0.0}$&  $4.634$& $0.9537$& $21.0$&$14.0$& ${0.0}$& $0.0063$& $0.0701$ \\

    & & & & \taggen & $12, 480$ & {$79.356$}& {$0.1806$}&  {$17.0$}&  {$\textbf{0.0}$}& {$0.7732$}& {$\textbf{0.0062}$}&{$6.5$}& {$\textbf{1.0}$}& {$\textbf{0.0}$}& {$\textbf{0.0009}$}& {$\textbf{0.0067}$} \\ 
    
     & & & & \namemodel & $\textbf{125}$ & {$\textbf{25.0}$}& {$\textbf{0.076}$}&  {$\textbf{12.0}$}&  {$\textbf{0.0}$}& {$\textbf{0.4135}$}& {$0.0102$}&{$\textbf{3.0}$}& {$3.0$}& {$\textbf{0.0}$}& {$0.0019$}&  {$0.013$} \\ 
     
\cmidrule(lr){1-17}

    \multirow{3}{1.2cm}{Bitcoin} & \multirow{3}{0.9cm}{$3.7K$}& \multirow{3}{0.9cm}{$24K$}&\multirow{3}{0.8cm}{ $191$}& \textit{Median}& - & - & $1.8443$&  $189.5$&  $2.0$& $3.5607$& $0.941$&$50.5$& $13.0$& $0.0102$& $0.0146$& $0.1003$\\
    
     & & & & \taggen & $18579$ & {$80.0$}& {$0.2311$}&  {$\textbf{23.0}$}& {$\textbf{0.0}$}& {$0.5207$}& {$\textbf{0.0045}$}&{$13.0$}& {$\textbf{1.0}$}& {$\textbf{0.0016}$}& {$\textbf{0.002}$}&  {$\textbf{0.0133}$} \\ 
     
     & & & & \namemodel & $\textbf{128}$ & {$\textbf{24.294}$}& {$\textbf{0.1217}$}&  {$25.5$}& {$1.0$}&{$\textbf{0.294}$}& {$0.0081$}& {$\textbf{6.0}$}&{$3.0$}& {$0.0078$}& {$0.0066$}&  {$0.0229$}\\ 

\cmidrule(lr){1-17}

      \multirow{2}{1.2cm}{ Wiki} & \multirow{2}{0.9cm}{$9.2K$}& \multirow{2}{0.9cm}{$157K$}&\multirow{2}{0.8cm}{$2.6M$}& \textit{Median}& - &- &$1.1525$& $33.0$&  $0.0$&$12.0371$& $0.9868$& $7.0$& $62.5$& $0.0$& $0.0$& $0.0093$\\
       &  &  &  & \namemodel  &{$896$} & {$25.573$}& {$0.072$}&  {$14.0$}& {$0.0$}&{$9.4139$}& {$0.0057$}&{$2.0$}& {$10.0$}& {$0.0$}& {$0.0$}& {$0.0017$}\\ 
    \cmidrule(lr){1-17}

       \multirow{2}{1.2cm}{ Reddit} & \multirow{2}{0.9cm}{$10.9K$}& \multirow{2}{0.9cm}{$662K$}&\multirow{2}{0.8cm}{$2.6M$}& \textit{Median}& - &- &$1.6693$& $5955.0$&  $0.0$& $5.672$&$0.916$& $269.0$& $147.0$& $0.0$& $0.0008$& $0.0217$ \\   
        
         &  &  & &  \namemodel & {$4661$}  &   {$0.077$}& {$0.1206$}&  {$704.0$}&  {$0.0$}&{$1.5709$}& {$0.0043$}& {$128.0$}&{$54.0$}& {$0.0$}& {$0.0006$}&  {$0.0099$} \\
        \cmidrule(lr){1-17}
       \multirow{2}{1.2cm}{Ta-feng} & \multirow{2}{0.9cm}{$56K$}& \multirow{2}{0.9cm}{$817K$}&\multirow{2}{0.8cm}{$120$}&  \textit{Median} &- & - & $2.8832$&  $52477.0$&  $0.0$& $2.6827$&$0.9289$& $3798.0$&$75.0$& $0.0$& $0.0011$& $0.1526$ \\
       
       & & & & \namemodel& {$2722$} & {$17.028$}& {$0.301$}&  {$14779.0$}& {$0.0$}& {$0.0762$}& {$0.0147$}&{$282.0$}& {$295.0$}& {$0.0$}& {$0.0003$}& {$0.045$}\\
        
        \bottomrule
\end{tabular}
}
\vspace{-0.10in}
    \caption{{\namemodel's performance against \taggen and \dymond in terms of graph generation time (Col 6), edge duplication percent (Col 7), and median error across various graph statistics (Cols 8-17). For all performance metrics, lower values are better. For each statistic, we also list the \textit{Median} value over \emph{original graph snapshots} to better contextualize the error values. The best result in each dataset is in boldface. We do not report the results for an algorithm if it does not complete within 24 hours. Errors smaller than five decimal places are approximated to $0$.}}
    \label{tab:result}
    \vspace{-0.20in}
\end{table*}

\begin{table}
    \centering
    \resizebox{0.8\columnwidth}{!}{
        \begin{tabular}{llll}
\toprule
        \textbf{Metric} & \textbf{Wiki-Small} & \textbf{UC Irvine} & \textbf{Bitcoin} \\
        \midrule
        \textbf{Generation Time(sec)} & $19$ & $1112$ & $640$ \\
        \textbf{$\%$Edge overlap} & $0$ & $0$ & $0$ \\
        \cmidrule(lr){1-4}
        \textbf{Mean degree} & $0.0463/1.1064$ & $0.1949/1.5714$ & $0.4015/1.8443$ \\
        \textbf{Wedge Count} & $7.0/13.0$ & $26.0/71.0$ & $50.0/190$ \\
        \textbf{Triangle Count} & $0.0/0.0$ & $0.0/0.0$ & $1.0/2.0$\\
        \textbf{PLE} & $5.3916/16.4626$ & $1.1036/4.634$ & $2.0456/3.5607$ \\
        \textbf{Edge Entropy} & $0.0045/.9912$ & $0.0125/.9537$ & $0.0177/.941$ \\
        \textbf{LCC} & $1.0/5$ & $7.0/21.0$ & $16.0/50.5$ \\
        \textbf{NC} & $4.0/44.0$ & $7.0/14.0$ & $18.0/13.0$ \\
        \textbf{Global CF} & $0.0/0.0$ &$0.0/0.0$ &$0.0096/0.0102$ \\
        \textbf{Mean BC} &$0.0/0.0$ & $0.0026/0.0063$ & $0.0113/0.0146$ \\
        \textbf{Mean CC} & $0.0015/0.0122$ & $0.0257/0.0701$ & $0.0594/0.1003$ \\
        \bottomrule
        \end{tabular}
        }
        \vspace{-0.10in}
    \caption{Median errors across various graph statistics for inductive version. Each entry, row 3 onwards, denotes $($median absolute error/ median value of the corresponding original graph property across snapshots$)$.}
    \vspace{-0.20in}
    \label{tab:inductive_result}
\end{table}

In this section, we benchmark \namemodel against \dymond and \taggen and establish that it \textbf{(1)} it is up to 2000 times faster, \textbf{(2)} breaks new ground on scalability against number of timestamps, and \textbf{(3)} generates graphs of high fidelity. Our codebase and datasets are available at  \url{https://github.com/data-iitd/tigger}.
\vspace{-0.05in}
\subsection{Experimental Setup}

\noindent
\textbf{Datasets:} For our empirical evaluation, we use the publicly available datasets listed in Table \ref{tab:result}. Columns 2 to 4 of Table \ref{tab:result} summarize the sizes of the temporal interaction graphs. Our datasets span various domains including message exchange platform (\textit{\bfseries UC Irvine})~\cite{konect}, financial network (\textit{\bfseries Bitcoin})~\cite{kumar2016edge}, communication forum (\textit{\bfseries Reddit})~\cite{snapnets}, shopping (\textit{\bfseries Ta-feng})~\cite{10.1145/3209978.3210129}, and Wikipedia edits (\textit{\bfseries Wiki})~\cite{snapnets}. Further details are provided in Table  \ref{tab:urls_datasets} in appendix.  Since \dymond and \taggen do not scale to graphs with large number of timestamps, we sample a smaller subset of Wiki by considering only the first 50 hours. This dataset is denoted as \textit{\bfseries Wiki-Small}.
\noindent
\textbf{Baselines and Training:} We benchmark the performance of \namemodel against 
\dymond and \taggen. For \namemodel, we denote the inductive version as \namemodel-I. To allow uniform comparison, in \namemodel-I, we generate graph of the same size as the source. For both \taggen and \dymond, we use the code shared by authors. For all algorithms, the entire input graph is used for training and a single synthetic graph is generated. Parameter details along with machine configuration are provided in the appendix.

\noindent
\textbf{Evaluation metrics:} 
The performance of a generative model is satisfactory if \textbf{(1)} it runs fast, \textbf{(2)} generates graphs with similar properties as in the source, \textbf{(3)} but without duplicating the source itself. To quantify these three objectives, we utilize the following metrics. 
\begin{itemize}
\item \textbf{Efficiency:} Efficiency is measured through running time of the graph generation component. 
\item \textbf{Fidelity:} To quantify preservation of original graph properties, we compare various graph statistics of the \textit{snapshots} of original graph  $\mathcal{G}_t$ and synthetic graph ${\mathcal{G'}}_{t}$ for each unique timestamp $t \in \{1 \ldots T\}$. We use the following graph statistics~\cite{kunegis2017handbook}: \textbf{\textit{(i)}} mean degree, \textbf{\textit{(ii)}} wedge count, \textbf{\textit{(iii)}} triangle count, \textbf{\textit{(iv)}} power law exponent of degree distribution \textit{(PLE)}, \textbf{\textit{(v)}} relative edge distribution entropy,  \textbf{\textit{(vi)}} largest connected component size \textit{(LCC)}, \textbf{\textit{(vii)}} number of components \textit{(NC)}, \textbf{\textit{(viii)}} global clustering coefficient \textit{(CF)}, \textbf{\textit{(ix)}} mean betweenness centrality \textit{(BC)}, \textbf{\textit{(x)}} mean closeness centrality \textit{(CC)}. We explain these metrics in Table \ref{tab:metrices_description} in appendix. The \emph{error} with respect to a given graph statistic $P$ is quantified as the \textit{median absolute error}, that is, $Median_{t\in[1\ldots T]}\lvert P(\CG_t)-P(\CG'_t)\rvert$. We use median instead of mean to reduce the impact of outliers. Nonetheless, the mean absolute errors (MAE) are also reported in the appendix.
\item \textbf{Duplication:} To capture the level of duplication, we compute the percentage of overlapping edges, i.e., $\frac{\lvert\CE\cap\CE'\rvert}{\lvert\CE\rvert}\times 100$. Measuring duplication is important since an algorithm that duplicates the source graph would obtain perfect scores with respect to property preservation, although the generated graph is of limited use. We note that duplication has not been studied by \taggen or \dymond. 
\end{itemize}
\begin{figure}[b]
\centering
\vspace{-0.20in}
\includegraphics[width=0.75\columnwidth]{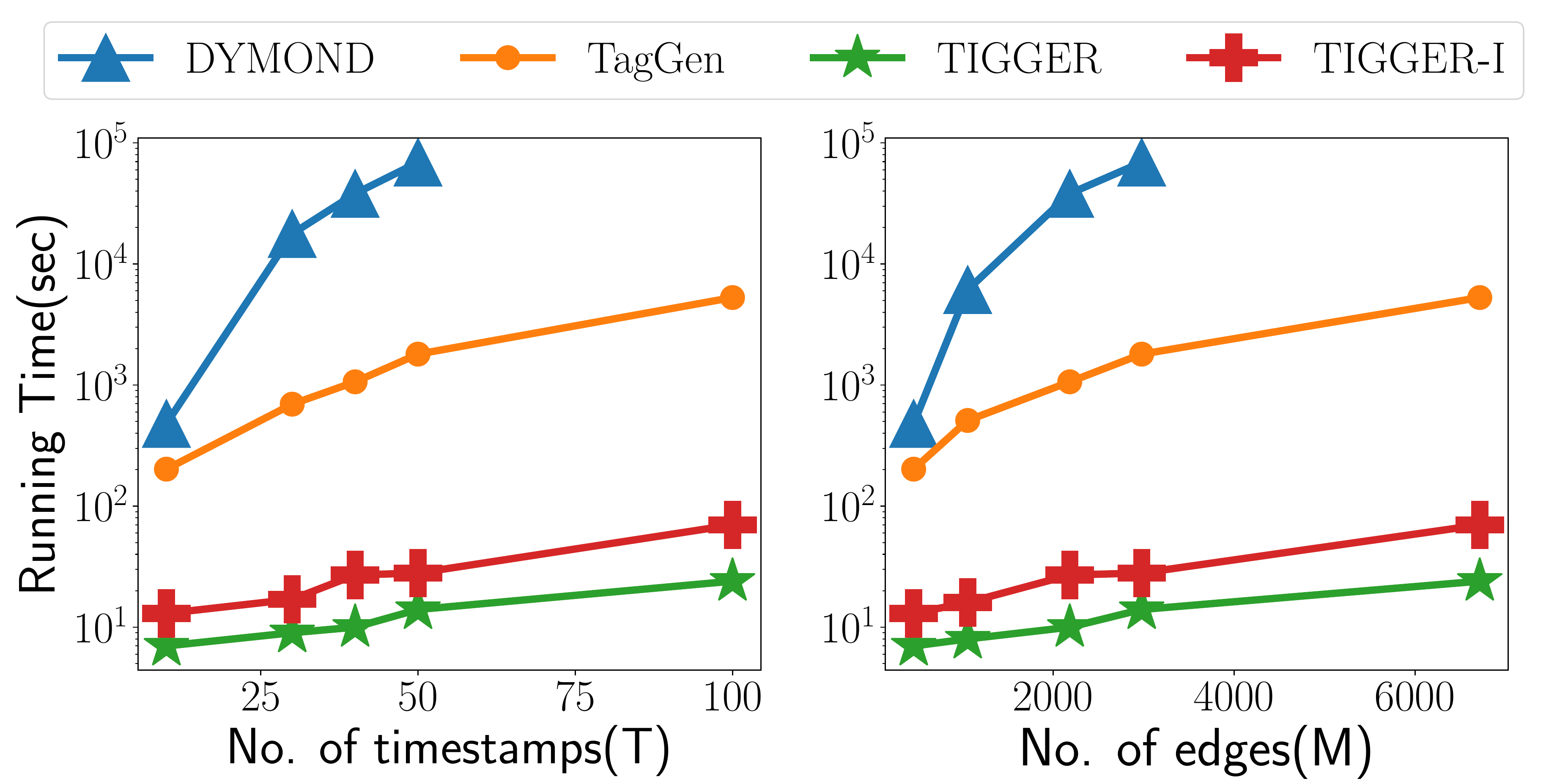}
\vspace{-0.10in}
\caption{{Scalability against  \# timestamps and \# edges in Wiki(hourly). The $y$-axis is in log-scale.}
\label{fig:timestamptime}
}
\vspace{-0.25in}
\end{figure}
\vspace{-0.1in}
\subsection{Transductive: Comparison against Baselines}

Table~\ref{tab:result} presents the performance of all transductive algorithms across all metrics. We summarize the key observations below.

\noindent
\textbf{Efficiency and Scalability:} \namemodel is by far the most efficient of all models, while \dymond is the slowest due to its $O(N^3T)$ time complexity. \taggen is nearly 2 orders slower than \namemodel. In the inference phase, \taggen samples paths from the original graph, uses heuristics to modify these paths and then employs a discriminator to select from the generated paths. This process is prohibitively slow. Additionally, \taggen performs an expensive inversion of $N'\times N'$ matrix where $N'$ is number of unique pairs of nodes and their interaction timestamps in $\mathcal{G}$. Consequently, \taggen fails to scale on Wiki and Reddit with millions of timestamps, and on Ta-feng which has much larger node and edge sets (see Table~\ref{tab:result}). Note that \dymond fails to complete in all but Wiki-Small, the smallest dataset. 
\namemodel on the other hand is orders of magnitude faster, and can scale to large datasets, since it simply uses the trained RNN to sample paths, and generates the graph using these paths. 
In Fig.~\ref{fig:timestamptime}, we plot the growth of running time against the number of timestamps and graph size. As visible, \namemodel is not only faster, but also have a slower growth rate. For this experiment, we sample the desired number of timestamps/temporal edges from the Wiki dataset.


\noindent
\textbf{Duplication:} \taggen consistently duplicates $\approx 80\%$ of the original graph. Hence, the utility of \taggen as a graph generator is questionable. Both \dymond and \namemodel do not suffer from this limitation.

\noindent
\textbf{Fidelity:} From Table~\ref{tab:result}, we observe that \namemodel and \taggen achieve the best results in majority of graph statistics. However, as we noted above, \taggen nearly duplicates the original graph and hence it is not surprising that the graph statistics remain nearly the same. In contrast, \namemodel has an edge overlap of $\approx 20\%$ on average, and yet achieves low errors similar to a near-duplicate graph. While \namemodel-I exhibits higher median error than the transductive \namemodel, it is better than \dymond (See Wiki-Small in Table~\ref{tab:inductive_result}). 

\begin{figure}[b]
\centering
\vspace{-0.20in}
\includegraphics[width=3.3in]{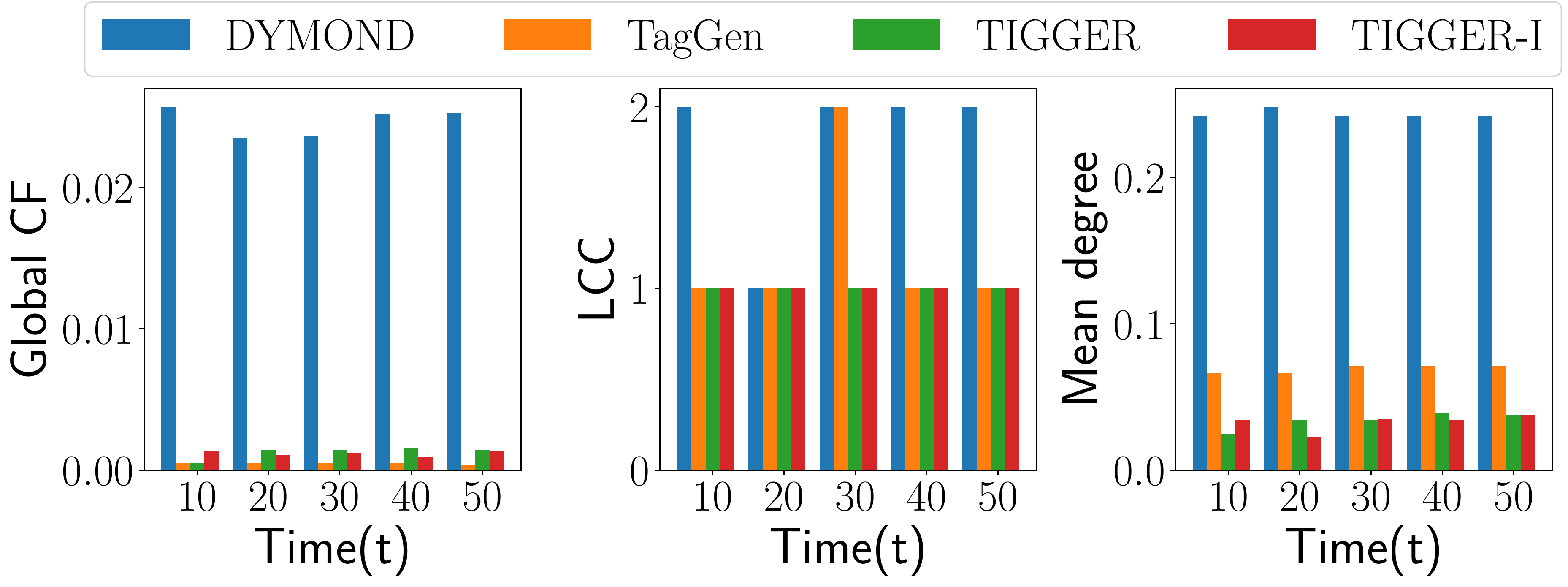}
\vspace{-0.25in}
\caption{{Median error over each consecutive window of 10 graph snapshots in Wiki-Small.}}
\label{fig:timestamped_metrics}
\vspace{-0.25in}
\end{figure}

To study how the performance varies with growth of graphs, we study the variation of median error against time in  Fig.~\ref{fig:timestamped_metrics}. Consistent with the trends in Table~\ref{tab:result}, the performance of \dymond is the weakest. \taggen performs marginally better in clustering coefficient (CF), while \namemodel is superior in mean degree and LCC. 
\vspace{-0.05in}
\subsection{Inductive: Performance of \namemodel-I}
Before initiating the discussion, we note that \namemodel-I offers an important feature not found in any of the transductive models, \textit{viz.}, the ability to control the size of the generated 
graph. Table~\ref{tab:inductive_result} presents the results.

\noindent
\textbf{Scalability:} \namemodel-I, is orders of magnitude faster than \taggen and \dymond. However, \namemodel-I is $5$--$8$ times slower than  transductive \namemodel. This is unsurprising since \namemodel-I needs to perform nearest neighbor search in the inference phase on node embeddings. Additionally, \namemodel-I is challenging to train on large graphs due to its reliance on \textsc{Wgan}, which often fails to converge on large graphs. Hence, we have not reported results on full Wiki, Reddit and Bitcoin. Fig.~\ref{fig:timestamptime} reveals that the growth rate of running time in \namemodel-I is similar to \namemodel.

\noindent
\textbf{Duplication:} The edge-overlap of \namemodel-I is 0 across all benchmarked datasets, which is the ideal score.

\noindent
\textbf{Fidelity:} The modelling task in inductive mode is inherently more difficult due to not having access to node IDs. Despite this challenge, we observe that the errors are low when compared to the median values of graph statistics in the original graph (Table~\ref{tab:inductive_result}). More importantly, despite being inductive, the errors are significantly better than \dymond and comparable to \taggen and \namemodel (compare Tables~\ref{tab:inductive_result} and \ref{tab:result}). This trend is also visible in Fig.~\ref{fig:timestamped_metrics}.

\vspace{-0.10in}
\section{Conclusion}
The success of a temporal graph generative model rests on two key properties: (1) Scalability to large temporal graphs since real-world graphs are large, and (2) and the ability to learn the underlying distribution of rules governing graph evolution rather than duplicating the training graph. Existing techniques fail to show the above desired behaviour. As established in our empirical evaluation, the proposed method, \namemodel, achieves the above desiderata. \namemodel derives its power through an innovative use of \textit{intensity-free temporal point processes} to jointly model the node interaction times and the structural properties of the source graph. 
Additionally, we introduce an inductive version called \namemodel-I, which directly learns the distribution over node embeddings instead of node IDs. 
\textbf{Future Work:} The scalability of the inductive model is limited by its graph embeddings and the use of \textsc{Wgan}. Hence, we plan to explore mechanisms that address this limitation and eventually move towards a model that is inductive, scalable and accurate in terms of fidelity.



\bibliography{references}

\begin{thebibliography}{38}
\providecommand{\natexlab}[1]{#1}

\bibitem[{Albert and Barab{\'a}si(2002)}]{albert2002statistical}
Albert, R.; and Barab{\'a}si, A.-L. 2002.
\newblock Statistical mechanics of complex networks.
\newblock \emph{Reviews of modern physics}, 74(1): 47.

\bibitem[{Arjovsky, Chintala, and Bottou(2017)}]{wgan}
Arjovsky, M.; Chintala, S.; and Bottou, L. 2017.
\newblock {W}asserstein Generative Adversarial Networks.
\newblock In Precup, D.; and Teh, Y.~W., eds., \emph{Proceedings of the 34th
  International Conference on Machine Learning}, volume~70 of \emph{Proceedings
  of Machine Learning Research}, 214--223. PMLR.

\bibitem[{Bai et~al.(2018)Bai, Nie, Zhao, Zhu, Du, and
  Wen}]{10.1145/3209978.3210129}
Bai, T.; Nie, J.-Y.; Zhao, W.~X.; Zhu, Y.; Du, P.; and Wen, J.-R. 2018.
\newblock \emph{An Attribute-Aware Neural Attentive Model for Next Basket
  Recommendation}, 1201–1204.
\newblock New York, NY, USA: Association for Computing Machinery.
\newblock ISBN 9781450356572.

\bibitem[{Bojchevski et~al.(2018)Bojchevski, Shchur, Z{\"u}gner, and
  G{\"u}nnemann}]{pmlr-v80-bojchevski18a}
Bojchevski, A.; Shchur, O.; Z{\"u}gner, D.; and G{\"u}nnemann, S. 2018.
\newblock {N}et{GAN}: Generating Graphs via Random Walks.
\newblock In Dy, J.; and Krause, A., eds., \emph{Proceedings of the 35th
  International Conference on Machine Learning}, volume~80 of \emph{Proceedings
  of Machine Learning Research}, 610--619. PMLR.

\bibitem[{Casas-Roma, Herrera-Joancomart\'{\i}, and Torra(2017)}]{dataprivacy}
Casas-Roma, J.; Herrera-Joancomart\'{\i}, J.; and Torra, V. 2017.
\newblock A Survey of Graph-Modification Techniques for Privacy-Preserving on
  Networks.
\newblock \emph{Artif. Intell. Rev.}, 47(3): 341–366.

\bibitem[{Dal~Pozzolo et~al.(2018)Dal~Pozzolo, Boracchi, Caelen, Alippi, and
  Bontempi}]{8038008}
Dal~Pozzolo, A.; Boracchi, G.; Caelen, O.; Alippi, C.; and Bontempi, G. 2018.
\newblock Credit Card Fraud Detection: A Realistic Modeling and a Novel
  Learning Strategy.
\newblock \emph{IEEE Transactions on Neural Networks and Learning Systems},
  29(8): 3784--3797.

\bibitem[{De~Cao and Kipf(2018)}]{cao2018molgan}
De~Cao, N.; and Kipf, T. 2018.
\newblock {MolGAN: An implicit generative model for small molecular graphs}.
\newblock \emph{ICML 2018 workshop on Theoretical Foundations and Applications
  of Deep Generative Models}.

\bibitem[{Goyal, Jain, and Ranu(2020)}]{goyal2020graphgen}
Goyal, N.; Jain, H.~V.; and Ranu, S. 2020.
\newblock GraphGen: a scalable approach to domain-agnostic labeled graph
  generation.
\newblock In \emph{Proceedings of The Web Conference 2020}, 1253--1263.

\bibitem[{Guo et~al.(2020)Guo, Sun, Lindgren, Geng, Simcha, Chern, and
  Kumar}]{pmlr-v119-guo20h}
Guo, R.; Sun, P.; Lindgren, E.; Geng, Q.; Simcha, D.; Chern, F.; and Kumar, S.
  2020.
\newblock Accelerating Large-Scale Inference with Anisotropic Vector
  Quantization.
\newblock In III, H.~D.; and Singh, A., eds., \emph{Proceedings of the 37th
  International Conference on Machine Learning}, volume 119 of
  \emph{Proceedings of Machine Learning Research}, 3887--3896. PMLR.

\bibitem[{Hamilton, Ying, and Leskovec(2017)}]{hamilton2018inductive}
Hamilton, W.~L.; Ying, R.; and Leskovec, J. 2017.
\newblock Inductive Representation Learning on Large Graphs.
\newblock In \emph{Proceedings of the 31st International Conference on Neural
  Information Processing Systems}, NIPS'17, 1025–1035. Red Hook, NY, USA:
  Curran Associates Inc.
\newblock ISBN 9781510860964.

\bibitem[{Han, Pei, and Kamber(2011)}]{han2011data}
Han, J.; Pei, J.; and Kamber, M. 2011.
\newblock \emph{Data mining: concepts and techniques}.
\newblock Elsevier.

\bibitem[{He and McAuley(2016)}]{amazon_dataset}
He, R.; and McAuley, J. 2016.
\newblock Ups and Downs: Modeling the Visual Evolution of Fashion Trends with
  One-Class Collaborative Filtering.
\newblock In \emph{Proceedings of the 25th International Conference on World
  Wide Web}, WWW '16, 507–517. Republic and Canton of Geneva, CHE:
  International World Wide Web Conferences Steering Committee.
\newblock ISBN 9781450341431.

\bibitem[{Hrinchuk, Popova, and Ginsburg(2020)}]{popova2019molecularrnn}
Hrinchuk, O.; Popova, M.; and Ginsburg, B. 2020.
\newblock Correction of Automatic Speech Recognition with Transformer
  Sequence-To-Sequence Model.
\newblock In \emph{2020 {IEEE} International Conference on Acoustics, Speech
  and Signal Processing, {ICASSP} 2020, Barcelona, Spain, May 4-8, 2020},
  7074--7078. {IEEE}.

\bibitem[{Ji et~al.(2021)Ji, Huang, Chen, and Xi}]{ji2021generating}
Ji, Y.; Huang, R.; Chen, J.; and Xi, Y. 2021.
\newblock Generating a Doppelganger Graph: Resembling but Distinct.
\newblock \emph{ArXiv}, abs/2101.09593.

\bibitem[{Karo{\'{n}}ski and Ruci{\'{n}}ski(1997)}]{Karonski1997}
Karo{\'{n}}ski, M.; and Ruci{\'{n}}ski, A. 1997.
\newblock \emph{The Origins of the Theory of Random Graphs}, 311--336.
\newblock Berlin, Heidelberg: Springer Berlin Heidelberg.
\newblock ISBN 978-3-642-60408-9.

\bibitem[{Kazemi et~al.(2019)Kazemi, Goel, Eghbali, Ramanan, Sahota, Thakur,
  Wu, Smyth, Poupart, and Brubaker}]{kazemi2019time2vec}
Kazemi, S.~M.; Goel, R.; Eghbali, S.; Ramanan, J.; Sahota, J.; Thakur, S.; Wu,
  S.; Smyth, C.; Poupart, P.; and Brubaker, M.~A. 2019.
\newblock Time2Vec: Learning a Vector Representation of Time.
\newblock \emph{ArXiv}, abs/1907.05321.

\bibitem[{Kingma and Welling(2014)}]{kingma2014autoencoding}
Kingma, D.~P.; and Welling, M. 2014.
\newblock Auto-Encoding Variational Bayes.
\newblock In Bengio, Y.; and LeCun, Y., eds., \emph{2nd International
  Conference on Learning Representations, {ICLR} 2014, Banff, AB, Canada, April
  14-16, 2014, Conference Track Proceedings}.

\bibitem[{Kumar et~al.(2016)Kumar, Spezzano, Subrahmanian, and
  Faloutsos}]{kumar2016edge}
Kumar, S.; Spezzano, F.; Subrahmanian, V.; and Faloutsos, C. 2016.
\newblock Edge weight prediction in weighted signed networks.
\newblock In \emph{Data Mining (ICDM), 2016 IEEE 16th International Conference
  on}, 221--230. IEEE.

\bibitem[{Kunegis(2013{\natexlab{a}})}]{konect}
Kunegis, J. 2013{\natexlab{a}}.
\newblock KONECT: The Koblenz Network Collection.
\newblock In \emph{Proceedings of the 22nd International Conference on World
  Wide Web}, WWW '13 Companion, 1343–1350. New York, NY, USA: Association for
  Computing Machinery.
\newblock ISBN 9781450320382.

\bibitem[{Kunegis(2013{\natexlab{b}})}]{kunegis2017handbook}
Kunegis, J. 2013{\natexlab{b}}.
\newblock KONECT: The Koblenz Network Collection.
\newblock In \emph{Proceedings of the 22nd International Conference on World
  Wide Web}, WWW '13 Companion, 1343–1350. New York, NY, USA: Association for
  Computing Machinery.
\newblock ISBN 9781450320382.

\bibitem[{Leskovec and Krevl(2014)}]{snapnets}
Leskovec, J.; and Krevl, A. 2014.
\newblock {SNAP Datasets}: {Stanford} Large Network Dataset Collection.
\newblock \url{http://snap.stanford.edu/data}.

\bibitem[{Li, Zhang, and Liu(2018)}]{li2018multi}
Li, Y.; Zhang, L.; and Liu, Z. 2018.
\newblock Multi-objective de novo drug design with conditional graph generative
  model.
\newblock \emph{Journal of cheminformatics}, 10(1): 1--24.

\bibitem[{Liao et~al.(2019)Liao, Li, Song, Wang, Hamilton, Duvenaud, Urtasun,
  and Zemel}]{liao2019gran}
Liao, R.; Li, Y.; Song, Y.; Wang, S.; Hamilton, W.~L.; Duvenaud, D.; Urtasun,
  R.; and Zemel, R.~S. 2019.
\newblock Efficient Graph Generation with Graph Recurrent Attention Networks.
\newblock In Wallach, H.~M.; Larochelle, H.; Beygelzimer, A.;
  d'Alch{\'{e}}{-}Buc, F.; Fox, E.~B.; and Garnett, R., eds., \emph{Advances in
  Neural Information Processing Systems 32: Annual Conference on Neural
  Information Processing Systems 2019, NeurIPS 2019, December 8-14, 2019,
  Vancouver, BC, Canada}, 4257--4267.

\bibitem[{Liu, Benson, and Charikar(2019)}]{Liu-2019-sampling}
Liu, P.; Benson, A.~R.; and Charikar, M. 2019.
\newblock Sampling methods for counting temporal motifs.
\newblock In \emph{Proceedings of the ACM International Conference on Web
  Search and Data Mining}.

\bibitem[{Mei and Eisner(2017)}]{mei2017neural}
Mei, H.; and Eisner, J. 2017.
\newblock The Neural Hawkes Process: A Neurally Self-Modulating Multivariate
  Point Process.
\newblock In \emph{Proceedings of the 31st International Conference on Neural
  Information Processing Systems}, NIPS'17, 6757–6767. Red Hook, NY, USA:
  Curran Associates Inc.
\newblock ISBN 9781510860964.

\bibitem[{Michail(2015)}]{Michail2015}
Michail, O. 2015.
\newblock \emph{An Introduction to Temporal Graphs: An Algorithmic
  Perspective}, 308--343.
\newblock Cham: Springer International Publishing.
\newblock ISBN 978-3-319-24024-4.

\bibitem[{Omi, Ueda, and Aihara(2019)}]{omi2020fully}
Omi, T.; Ueda, N.; and Aihara, K. 2019.
\newblock Fully Neural Network based Model for General Temporal Point
  Processes.
\newblock In Wallach, H.~M.; Larochelle, H.; Beygelzimer, A.;
  d'Alch{\'{e}}{-}Buc, F.; Fox, E.~B.; and Garnett, R., eds., \emph{Advances in
  Neural Information Processing Systems 32: Annual Conference on Neural
  Information Processing Systems 2019, NeurIPS 2019, December 8-14, 2019,
  Vancouver, BC, Canada}, 2120--2129.

\bibitem[{Paranjape, Benson, and Leskovec(2017)}]{Paranjape_2017}
Paranjape, A.; Benson, A.~R.; and Leskovec, J. 2017.
\newblock Motifs in Temporal Networks.
\newblock \emph{Proceedings of the Tenth ACM International Conference on Web
  Search and Data Mining}.

\bibitem[{Ranu and Singh(2009)}]{ranu2009graphsig}
Ranu, S.; and Singh, A.~K. 2009.
\newblock Graphsig: A scalable approach to mining significant subgraphs in
  large graph databases.
\newblock In \emph{2009 IEEE 25th International Conference on Data
  Engineering}, 844--855. IEEE.

\bibitem[{Rizoiu et~al.(2017)Rizoiu, Lee, Mishra, and Xie}]{rizoiu2017tutorial}
Rizoiu, M.; Lee, Y.; Mishra, S.; and Xie, L. 2017.
\newblock A Tutorial on Hawkes Processes for Events in Social Media.
\newblock \emph{CoRR}, abs/1708.06401.

\bibitem[{Shchur, Bilo\v{s}, and G\"{u}nnemann(2020)}]{shchur2020intensityfree}
Shchur, O.; Bilo\v{s}, M.; and G\"{u}nnemann, S. 2020.
\newblock Intensity-Free Learning of Temporal Point Processes.
\newblock \emph{International Conference on Learning Representations (ICLR)}.

\bibitem[{Vaswani et~al.(2017)Vaswani, Shazeer, Parmar, Uszkoreit, Jones,
  Gomez, Kaiser, and Polosukhin}]{vaswani2017attention}
Vaswani, A.; Shazeer, N.; Parmar, N.; Uszkoreit, J.; Jones, L.; Gomez, A.~N.;
  Kaiser, L.; and Polosukhin, I. 2017.
\newblock Attention Is All You Need.
\newblock \emph{CoRR}, abs/1706.03762.

\bibitem[{Walker(1977)}]{10.1145/355744.355749}
Walker, A.~J. 1977.
\newblock An Efficient Method for Generating Discrete Random Variables with
  General Distributions.
\newblock \emph{ACM Trans. Math. Softw.}, 3(3): 253–256.

\bibitem[{Watts~DJ(1998)}]{small_world}
Watts~DJ, S.~S. 1998.
\newblock Collective dynamics of 'small-world' networks.
\newblock In \emph{Nature}.

\bibitem[{Yang et~al.(2013)Yang, Zhang, Yu, and Yu}]{foursquare}
Yang, D.; Zhang, D.; Yu, Z.; and Yu, Z. 2013.
\newblock Fine-Grained Preference-Aware Location Search Leveraging Crowdsourced
  Digital Footprints from LBSNs.
\newblock In \emph{Proceedings of the 2013 ACM International Joint Conference
  on Pervasive and Ubiquitous Computing}, UbiComp '13, 479–488. New York, NY,
  USA: Association for Computing Machinery.
\newblock ISBN 9781450317702.

\bibitem[{You et~al.(2018)You, Ying, Ren, Hamilton, and
  Leskovec}]{you2018graphrnn}
You, J.; Ying, R.; Ren, X.; Hamilton, W.; and Leskovec, J. 2018.
\newblock Graphrnn: Generating realistic graphs with deep auto-regressive
  models.
\newblock In \emph{International Conference on Machine Learning}, 5708--5717.
  PMLR.

\bibitem[{Zeno, La~Fond, and Neville(2021)}]{dymond}
Zeno, G.; La~Fond, T.; and Neville, J. 2021.
\newblock DYMOND: DYnamic MOtif-NoDes Network Generative Model.
\newblock In \emph{Proceedings of the Web Conference 2021}, WWW '21, 718–729.
  New York, NY, USA: Association for Computing Machinery.
\newblock ISBN 9781450383127.

\bibitem[{Zhou et~al.(2020)Zhou, Zheng, Han, and He}]{TagGen}
Zhou, D.; Zheng, L.; Han, J.; and He, J. 2020.
\newblock A Data-Driven Graph Generative Model for Temporal Interaction
  Networks.
\newblock In \emph{Proceedings of the 26th ACM SIGKDD International Conference
  on Knowledge Discovery; Data Mining}, KDD '20, 401–411. New York, NY, USA:
  Association for Computing Machinery.
\newblock ISBN 9781450379984.

\end{thebibliography}

\clearpage
\appendix
\begin{appendices}

\section{Code optimizations}
To increase the sampling efficiency during training, we restrict the $\mathcal{N}_t(v)$ by considering only the future $W$ edges. Moreover, we implement alias-table based sampling procedure \cite{10.1145/355744.355749} which enables $O(1)$ sampling. 

\section{Attention based $p(S)$}
$p(S)$ can be rewritten using attention\cite{vaswani2017attention} based generator too. In this case, the individual conditionals $p(s_i.v)$ and $p(s_i.t)$ will depend upon the whole sequence $\{s_1 \ldots s_{i-1}, s_{i+1} \ldots s_{\ell}\}$ and not only the hidden state. In order to perform future pair prediction with this model, future pairs need to be masked during training. Note that this is the same issue with bidirectional RNNs. Moreover, sequence length is less than 50 where LSTMs are known to perform well empirically. Hence, we are using LSTM as the $rnn_{\theta}$. 

\begin{table}[p]
    \centering
    \scalebox{0.62}{
        \begin{tabular}{m{0.25 \linewidth}  m{0.7 \linewidth}}
         \toprule
         \textbf{Symbol} & {\textbf{Meaning}} \\ 
         \midrule
         $\mathcal{G}$& Input temporal interaction graph \\ \cmidrule(lr){1-2}
         $\mathcal{G}'$& Synthetic temporal interaction graph \\ \cmidrule(lr){1-2}
         $\mathcal{V}$ & Set of nodes in $\mathcal{G}$ \\
         \cmidrule(lr){1-2}
         $M$ & Number of temporal edges in $\mathcal{G}$\\
         \cmidrule(lr){1-2}
         $N$ & Number of nodes in $\mathcal{G}$ \\
         \cmidrule(lr){1-2}
         $M'$ & Number of temporal edges in $\mathcal{G}'$\\
         \cmidrule(lr){1-2}
         $N'$ & Number of nodes in $\mathcal{G}'$ \\
         \cmidrule(lr){1-2}
         $e$& Edge tuple containing $(u,v,t)$ where $u$,$v \in \mathcal{V}$ and $t \in \mathcal{R}$ \\\cmidrule(lr){1-2}
         $\mathcal{E}$& Set of edge tuples $e$ \\
         \cmidrule(lr){1-2}
         $\mathcal{N}_{t}(v)$  & Temporal neighbourhood of a node $v$ at time $t$ \\ \cmidrule(lr){1-2}
          $\ell$  & Length of a temporal random walk \\ \cmidrule(lr){1-2}
          $s$  & Tuple containing node-time pair $(v, t)$  \\  \cmidrule(lr){1-2}
          $s.v$ & Node $v$ in the tuple $s$ \\\cmidrule(lr){1-2}
          $s.t$ & Time t in the tuple $s$ \\\cmidrule(lr){1-2}
          $\Delta t$ & Time difference between two consecutive tuples $s_i$ and $s_{i-1}$ of $S$  \\\cmidrule(lr){1-2}
          $S$  & Temporal random walk \\ \cmidrule(lr){1-2}
          $\mathcal{S}$ & Set of temporal random walks \\ \cmidrule(lr){1-2}
          $\ell'$  & Length of synthetic temporal random walk \\ \cmidrule(lr){1-2}
          $S'$  & Synthetic temporal random walk \\  \cmidrule(lr){1-2}
          $\mathcal{S}'$ & Set of synthetic temporal random walks \\ \cmidrule(lr){1-2}
          $\cf_v$ & Node id to vector transformation function \\ \cmidrule(lr){1-2}
          $\textbf{W}^v$ & Weights corresponding to $\cf_v$\\\cmidrule(lr){1-2}
          $\textbf{v}$ & Vector representation of node $v$ \\\cmidrule(lr){1-2}
          $\cf_{\textbf{v}}$ & Node embedding transformation function \\ \cmidrule(lr){1-2}
          $\textbf{W}$ & Weights corresponding to $\cf_{\textbf{v}}$\\\cmidrule(lr){1-2}
          $\cf_t$ & Time transformation function (Time2vec) \\ \cmidrule(lr){1-2}
          $rnn_{\theta}$ & An RNN cell parameterized by $\theta$ \\\cmidrule(lr){1-2}
          $\textbf{h}_i$ & Hidden state of $rnn_{\theta}$ at $i^{th}$ step  \\ \cmidrule(lr){1-2}
          $\co_i$ & Output of $rnn_{\theta}$  at $i^{th}$ step  \\ \cmidrule(lr){1-2}
          $\theta_v$ & Multinomial distribution over node $v$ parameterized in transductive recurrent generative model \\\cmidrule(lr){1-2}
          $\textbf{W}^O_v$ & Weights corresponding to $\theta_v$ \\\cmidrule(lr){1-2}
          $C$ & Number of components in \textit{log normal} mixture model \\ \cmidrule(lr){1-2}
          $\mu_c^{C} $ & Mean of $c^{th}$ component in \textit{log normal} mixture model \\ \cmidrule(lr){1-2}
          $\sigma_c^{C} $ & Std. dev of $c^{th}$ component in \textit{log normal} Mixture model \\ \cmidrule(lr){1-2}
          $\phi_c^{C}$ & Weightage of $c^{th}$ component in \textit{log normal} mixture model \\ \cmidrule(lr){1-2}
        $\textbf{W}^{{\mu C}}_c$ & Weights corresponding to $\mu_c^{C}$  \\\cmidrule(lr){1-2}
        $\textbf{W}^{{\sigma C}}_c$ & Weights corresponding to $\sigma_c^{C}$ \\ \cmidrule(lr){1-2}
        $\textbf{W}^{{\phi C}}_c$ & Weights corresponding to $\phi_c^{C}$\\ \cmidrule(lr){1-2}
        $K$ & Number of clusters in inductive recurrent generative model \\ \cmidrule(lr){1-2}
        $k_i$ & $i^{th}$ cluster in inductive recurrent generative model $\forall i \in \{1...K\}$ \\ \cmidrule(lr){1-2}
        $\boldsymbol{\mu}^{K}_{k_i}$, $\boldsymbol{\sigma}^{K}_{k_i}$ & Mean and std. deviation corresponding to normal distribution over $\textbf{z}$ given the cluster $k_i$ and $\co_i$ in inductive recurrent generative model\\ \cmidrule(lr){1-2}
        $\textbf{W}^{\mu K}_{ k_i}$, $\textbf{W}^{\sigma K}_{ k_i}$ & Weights corresponding to $\boldsymbol{\mu}^{K}_{k_i}$, $\boldsymbol{\sigma}^{K}_{k_i}$\\\cmidrule(lr){1-2}
        $\boldsymbol{z}$ & Latent variable introduced in inductive recurrent generative model \\\cmidrule(lr){1-2}
        $\boldsymbol{\mu}^{Z}$, $\boldsymbol{\sigma}^{Z}$ & Mean and std. deviation corresponding to normal distribution over $s_i.\textbf{v}$ given $\textbf{z}$ in inductive recurrent generative model\\ \cmidrule(lr){1-2}
        $\textbf{W}^{\mu Z}$, $\textbf{W}^{\sigma Z}$ & Weights corresponding to $\boldsymbol{\mu}^{Z}$, $\boldsymbol{\sigma}^{Z}$\\ \cmidrule(lr){1-2}
        $L$ & Number of samples to approximate the $\mathcal{E}$ term in inductive recurrent generative model \\\cmidrule(lr){1-2}
        $\mathcal{L}$ & Training loss \\ \bottomrule
        \end{tabular}
        }
    \captionsetup{justification=centering}
    \caption{Notations used in the paper}
    \label{tab:notation}
\end{table}

\section{Computational Complexity}

\textbf{Transductive}:

For constructing a temporal graph, we need to sample $O(M)$ temporal random walks. Each sampled temporal random walk is of length $\ell'$. Then, for each step we first pass the current node through a node embedding layer which is an $O(1)$ operation. Further, the time embedding layer of \textsc{Time2Vec} takes $O(d_T)$ time. The concatenated node and time embedding is passed through an LSTM with hidden layer size $d_H$ which takes $O(d_H \times (d_V + d_T))$ time. The mixture model has $C$ components, and computing $\mu_c^{C}$, $\sigma_c^{C}$ and $\phi_c^{C}$ for each component $c$ requires a dot product operation which is $O(d_V+d_O)$. Predicting the  next node is done by first computing a multinomial distribution over all nodes $v \in V$ and requires dot product operation of $O(d_O)$ which takes total time $O(N \times d_O)$.
Combining all terms, the time complexity of the transductive model is  $O ( M\times \ell' \times(N\times d_O  + C\times(d_V+d_O) + d_H\times(d_T+d_V)))$.
Ignoring the dimension terms of weight matrices, the above expression simplifies as $O ( M\times \ell' \times(N + C ))$.


\noindent
\textbf{Inductive:}
Similar to transductive variant,  we need to sample $O(M)$ temporal random walks for constructing a temporal graph. Each sampled temporal random walk is of length $\ell'$. Then, for each step, we first pass the current node through an MLP layer of hidden size $d_\textbf{V}$ which is $O(d_\textbf{V} \times d_\textbf{V} )$ operation. The time embedding layer of \textsc{Time2Vec} takes $O(d_T)$ time.  The concatenated node and time embedding is passed through an LSTM with hidden layer size $d_H$ which takes $O(d_H \times (d_\textbf{V} + d_T))$ time. Sampling the  next cluster requires computing multinomial distribution over $K$ clusters which takes total time $O(K (d_O))$.
Predicting $\boldsymbol{\mu}^K_k$ and  $\boldsymbol{\sigma}^K_k$  for each cluster $k \in K$ takes total time $O(K \times (d_O \times d_Z))$. The mixture model has $C$ components, and computing $\mu_c^{C}$, $\sigma_c^{C}$ and $\phi_c^{C}$ for each component $c$ requires $O(d_\textbf{V}+d_O)$ time. The number of nodes in the generated graph is $N'$. The time required to perform nearest neighbor search at each step is $O(N' \times d_V)$ where $d_\textbf{V}$ is the dimension of the sampled node embeddings from \textsc{WGAN}.

Combining all terms above, the time complexity of the inductive model is $O ( M\times \ell' \times( K\times (d_O \times d_Z) +  C\times(d_\textbf{V}+d_O)  + (d_\textbf{V})^2 + d_H\times(d_T+d_\textbf{V}) ) + N' \times d_\textbf{V})$.

As an optimization for the nearest neighbour search, we use ScANN \cite{pmlr-v119-guo20h} to perform faster approximate nearest neighbor search.
The running time complexity of ScANN is $O(qd+N')$ where $q$ is the size of each quantization codebook, $d$ is the dimension of vectors and $N'$ is the number of vectors. 

Simplifying the dimension terms as done earlier in transductive model, the total time complexity of inductive model \namemodel-I is $O ( M\times \ell' \times( K +  C  +  N'))$.



\section{Datasets and Pre-processing}
The semantics of the datasets are as follows:
\begin{itemize}
    \item \textbf{UC Irvine messages}: It is a homogeneous graph of messages exchange between students of UC Irvine~\cite{konect}. 
    \item \textbf{Bitcoin alpha network}: It is a homogeneous financial transaction graph of bitcoin trading between users of bitcoin-alpha trading platform~\cite{kumar2016edge}.
    \item \textbf{Reddit Interaction network}: Its a bipartite graph of users' post on subreddits ~\cite{snapnets}. In table \ref{tab:result}.
    \item \textbf{Wiki Edit}:  It a  bipartite graph between human editors and Wikipedia pages \cite{snapnets}. Additionally, we curate a small \textbf{Wiki- Small} which corresponds to first 50 hours of wiki edit. 
    \item \textbf{Ta-feng grocery shopping dataset}~\cite{10.1145/3209978.3210129}: It is a bipartite graph of grocery shopping dataset spanning from November 2000 to February 2001. 
\end{itemize}
\noindent
\textbf{Data Pre-processing:} Apart from removing the duplicate interactions at same timestamps, we don't perform any pre-processing on the dataset cited from the source.

\begin{table}[ht]
    \centering
    \scalebox{0.9}{
        \begin{tabular}{m{0.3\linewidth}  m{0.6 \linewidth}}\\
         \toprule
        \textbf{Baseline} & \textbf{Source}\\\midrule
        TagGen & \url{https://github.com/davidchouzdw/TagGen} \\    \cmidrule(lr){1-2}
        DYMOND & \url{https://github.com/zeno129/DYMOND}\\ \bottomrule
        \end{tabular}
        }
    \caption{Sources of baseline implementation}
    \label{tab:urls_code}
\end{table}
\begin{table}[ht]
    \centering
    \scalebox{0.9}{
        \begin{tabular}{m{0.3\linewidth}  m{0.6 \linewidth}} 
         \toprule
        \textbf{Dataset} & \textbf{Source}\\\midrule
        UC Irvine messages & \url{http://konect.cc/networks/opsahl-ucsocial/} \\    \cmidrule(lr){1-2}
        Bitcoin-alpha & \url{http://snap.stanford.edu/data/soc-sign-bitcoin-alpha.html}\\    \cmidrule(lr){1-2}
        Reddit Interaction network & \url{http://snap.stanford.edu/caw/} \\    \cmidrule(lr){1-2}
        Wiki Edit network & \url{http://snap.stanford.edu/caw/}\\    \cmidrule(lr){1-2}
        Ta-feng grocery shopping network & \url{https://www.kaggle.com/chiranjivdas09/ta-feng-grocery-dataset}, \url{ http://www.bigdatalab.ac.cn/benchmark/bm/dd?data=Ta-Feng}\\    \bottomrule
        \end{tabular}
        }
    \caption{Sources of datasets}
    \label{tab:urls_datasets}
\end{table}

\begin{table}[ht]
    \centering
    \scalebox{0.9}{
        \begin{tabular}{m{0.4 \linewidth}  m{0.5 \linewidth}} 
         \toprule
         \textbf{Metric} & \textbf{Description}\\ \midrule 
         
         Mean degree & Average of node degrees \\ 
         \cmidrule(lr){1-2}
        
           Wedge count & Number of two hops path  \\ \cmidrule(lr){1-2} 
      
        Triangle count & Number of triangles in the network \\  \cmidrule(lr){1-2} 
         Power law exponent(PLE) & Exponent of power law distribution on the node degrees  \\  \cmidrule(lr){1-2}
         Relative edge distribution entropy (RED entropy) & It measures the skewness of node degrees \\
         \cmidrule(lr){1-2} 
         Largest connected component size(LCC) & Size of largest connected component in the network  \\  \cmidrule(lr){1-2}
         Number of components(NC) & Number of connected component in the network \\  \cmidrule(lr){1-2}
         Global clustering coefficient(Global CF) & It is computed as the fraction of number of closed triplets and number of all triplets.\\
         \cmidrule(lr){1-2}
         Mean betweenness(BC)  & Mean of each node's betweeness centrality. Betweenness centrality of node $v$ is the fraction of all shortest paths which pass through $v$.\\
         \cmidrule(lr){1-2} 
         Mean Closeness centrality(CC)  & Mean of each node's closeness centrality. Closeness centrality of node is the reciprocal of average shortest path distance to other reachable nodes.\\  
          \bottomrule
        \end{tabular}
        }
    \caption{Description of undirected graph properties}
    \label{tab:metrices_description}
\end{table}

\section{Node representation using \sage}
For each node $v$ in the network $\mathcal{G}^{static}$, a representation $\textbf{v}$ is learnt by concatenating self information with information received from 1-hop neighbourhood by mean message passing. We utilize the following unsupervised loss on output representation $\textbf{v}$ to learn the message-passing parameters.
\begin{align}
\nonumber
    \mathcal{L} = -\log(\sigma(\textbf{v}^T\textbf{v}_j)) - Q\mathbb{E}_{ v_k \sim P_n(v)} \sigma(-\textbf{v}^T\textbf{v}_k)
    \label{eq:graphsageloss}
\end{align}
where $v_j \in \{u \mid d(u,v)=1\}$ and $Q$ is number of negative samples and $P_n(v)$ is probability distribution of negative nodes $v_k \in \{u \mid d(u,v) \neq 1\}$.  \cite{hamilton2018inductive}.
Please note that this method can produce  similar embeddings for multiple nodes even having no edges between them. Hence, we follow boosting training approach as suggested \cite{ji2021generating}. After 1 round of training, we increase of weight of nodes in $P_n(v)$ which contain false positive edge with node $v$. We repeat this process, until the number of false positive edges comes down below to certain threshold.

\section{\textsc{Wgan}}
We follow the similar training procedure as described in \cite{ji2021generating}. Given node embedding $\textbf v \; \forall v \in \mathcal{G}^{static}$, we initially remove the duplicate embeddings. Following this, we define a generator and  critic based on 3 layer MLP. Finally, we  optimize the \textsc{Wgan} value function by training generator for 1 epoch and critic for 4 epochs. We repeat this process until the convergence of loss. In order to avoid vanishing/explosion of gradients, we use \textsc{Wgan} along with gradient clipping. For training \textsc{Wgan} on \sage embeddings, we have used the code shared by \cite{ji2021generating}.

\begin{algorithm}
\caption{Sampling synthetic temporal random walks from a trained inductive recurrent generative model}
\label{alg:alg_ind}
{\scriptsize
\begin{algorithmic}[1]
 	 \REQUIRE  $S_1$,  ${\cf_{\textbf{v}},\cf_t,rnn_\theta,\theta_k ,\textbf{W}^{\mu K}_k, \textbf{W}^{\sigma K}_{k}\; \forall k \in \{1..K\},\theta_t},$${\textbf{W}^{\mu Z},\textbf{W}^{\sigma Z},\ell'}$
	 \ENSURE $\mathcal{S}'$
	 \STATE $\mathcal{S}' = \{\}$
	 \FOR{$s_1 \in S_1$}
	    \STATE $S' \gets \{\} ,(\textbf{v}_1,t_1 )\gets s_1$
	    \STATE $\textbf{v}_1 \gets \cf_{\textbf{v}}(\textbf{v}_1) , \textbf{t}_1 \gets \cf_t(t_1)$
	    \STATE $\ch_1 \gets \textbf{0}$
	    \FOR{$i \in \{2,3\ldots \ell' \}$}
	        \STATE $\co_i,\ch_i \gets rnn_{\theta}(\textbf{h}_{i-1},(\textbf{v}_{i-1} \concat \textbf{t}_{i-1}))$
	        \STATE $k_i \sim Multinomial(\theta_{k_1}(\co_i),\theta_{k_2}(\co_i) \ldots \theta_{k_K}(\co_i))$ \COMMENT{Sample next cluster} 
	        \STATE$\textbf{z} \sim \mathcal{N}(\textbf{W}^{\mu K}_{k_i}\co_i,\exp(\textbf{W}^{\sigma K}_{k_i}\co_i))$
	        \STATE$\textbf{v}_i \sim \mathcal{N}(\textbf{W}^{\mu Z}\textbf{z},\exp(\textbf{W}^{\sigma Z}\textbf{z})))$ \COMMENT{Sample next node embedding} 
	        \STATE $\Delta t \sim {\theta_t}(t-t_{i-1} \mid \textbf{v}_i,\co_i)$ \COMMENT{Sample next time using eq. \ref{eq:time_sample}}
	        \STATE $t_i = t_{i-1} +\Delta t$
	        \STATE $S'=S'+ (\textbf{v}_{i-1},\textbf{v}_i,t_i)$
	    \ENDFOR 
	 \STATE $\mathcal{S}' = \mathcal{S}'+ S'$
	 \ENDFOR
	 \STATE \textbf{Return} $\mathcal{S}'$
\end{algorithmic}
}
\end{algorithm}

\begin{table}[ht]
    \centering
    \resizebox{\columnwidth}{!}{
        \begin{tabular}{llll}
\toprule
        \textbf{Metric} & \textbf{Wiki-Small} & \textbf{UC Irvine} & \textbf{Bitcoin} \\
        \midrule
        \textbf{Mean degree} & $0.0701\pm 0.0665$ & $0.2589\pm 0.2734$ & $0.4302\pm 0.2697$ \\
        \textbf{Wedge Count} & $12.2449\pm 15.3377$ & $323.1081\pm 1099.1924$ & $147.2024\pm 354.5897$ \\
        \textbf{Triangle Count} & $0.0816\pm 0.3403$ & $2.9351\pm 7.6297$ & $4.2262\pm 11.1011$\\
        \textbf{PLE} & $7.0485\pm 6.6917$ & $2.0852\pm 3.8017$ & $3.6262\pm 5.8867$ \\
        \textbf{Edge Entropy} & $0.0062\pm 0.0053$ & $0.0191\pm 0.0243$ & $0.0222\pm 0.0191$ \\
        \textbf{LCC} & $3.0816\pm 4.5258$ & $14.4054\pm 18.8709$ & $21.9167\pm 19.9414$ \\
        \textbf{NC} & $6.898\pm 7.3436$ & $10.2486\pm 9.3988$ & $19.6786\pm 15.6461$ \\
        \textbf{Global CF} &$0.0066\pm 0.0272$ &$0.0338\pm 0.2473$ &$0.0273\pm 0.0577$ \\
        \textbf{Mean BC} &$0.0001\pm 0.0003$ & $0.0051\pm 0.0067$ & $0.0166\pm 0.0188$ \\
        \textbf{Mean CC} & $0.0025\pm 0.0028$ & $0.0388\pm 0.0437$ & $0.0718\pm 0.0556$ \\
        \bottomrule
        \end{tabular}
        }
        \vspace{-0.10in}
    \caption{Mean absolute errors across various graph statistics for inductive version. Each entry denotes $($Mean absolute error $\pm$ std. dev across snapshots$)$.}
    \vspace{-0.20in}
    \label{tab:inductive_result_mean}
\end{table}

\section{Training and Parameter details}
All experiments are performed on a machine running Intel Xeon E5-2698v4 processor with 64 cores,
having 1 Nvidia 1080 Ti GPU card with 11GB GPU memory, and 376 GB RAM running Ubuntu 16.04.

We set the length of a temporal random walk $(\ell)$ to 20 during training. We note that during training, we expanded the node set $\mathcal{V}$ by adding an additional node \textit{end\_node} to represent an empty temporal neighbourhood. We stop the generation of a temporal random walk if an \textit{end\_node} is sampled as the next node or max length is reached during sampling procedure.  We use 2 layer LSTM cell for $rnn_{\theta}$ and select $d_V=100$, $d_T=64$, $d_O=200$, $C=128$ and $K=300$. In \namemodel-I, we additionally set $d_\textbf{V}=128$ and $d_Z=128$. Both $d_\textbf{V}$ and $d_Z$ are constrained by the \sage embedding dimensions. To train both variants, we sample a single temporal random walk from every temporal edge of $\mathcal{G}$ thus collecting $M$ temporal random walks. We assume 1 training epoch as training over these $M$ temporal random walks. We re-sample $M$ temporal random walks from $\mathcal{G}$ for each succeeding round of epoch. We set $\mathcal{\beta}$ component of KL divergence term as $0.00001$. During graph generation, we set $\ell'$ as $\approx$ 2-5 for small graphs like wiki-edit and $\approx$ 6-10 for UC Irvine and Bitcoin networks. For both \dymond and \taggen we use the implementation provided by authors to learn the parameters from the input graph.

\section{Fidelity- Mean Errors}
In the main paper, we have reported the performance in terms of median absolute error. For the sake of completeness, we also report in Table \ref{tab:inductive_result_mean} and \ref{tab:result_mean} the Mean absolute error i.e. $Mean_{t\in[1\ldots T]}\lvert P(\CG_t)-P(\CG'_t)\rvert$ for all 5 datasets. Mean in the method column represent the mean of corresponding original graph statistic across timestamps. This is shown to represent the scale of the graph. Each value is in form of mean $\pm$ std. deviation.

\begin{table*}
\centering  
\resizebox{0.91\textwidth}{!}{

\begin{tabular}{lp{1.3cm}p{1.3cm}p{1.8cm}p{1.5cm}p{1.5cm}p{1.4cm}p{1.6cm}p{1.6cm}p{1.4cm}p{1.1cm}p{1.1cm}}
\toprule
    \textbf{Dataset} &  \textbf{Method} &  \textbf{Mean degree} &  \textbf{Wedge Count} &  \textbf{Triangle count} & \textbf{PLE} &\textbf{Edge entropy} & \textbf{LCC} & \textbf{NC} & \textbf{Global CF} & \textbf{Mean BC}   & \textbf{Mean CC}  \\ 
    \midrule
    
    \multirow{4}{1.2cm}{Wiki-Small}  &   \emph{{Mean}}  & $1.1131\pm 0.0427$& $19.5918\pm 17.1142$& $0.0\pm 0.0$& $17.7757\pm 7.3555$& $0.9885\pm 0.0071$& $5.9592\pm 2.5869$& $47.3469\pm 10.7163$& $0.0\pm 0.0$& $0.0001\pm 0.0$& $0.0125\pm 0.0028$ \\
    
    &  \dymond & $0.2419\pm 0.0495$& $11.2449\pm 14.0662$& $0.0\pm 0.0$& $12.6438\pm 7.3825$& $0.008\pm 0.0053$& $2.3878\pm 2.1459$& $33.4898\pm 11.4448$& $0.0\pm 0.0$& $0.0006\pm 0.0004$& $0.0281\pm 0.012$ \\ 

    & \taggen &  $0.0657\pm 0.0203$& $7.58\pm 4.0649$& $0.0\pm 0.0$& $6.7065\pm 5.1145$& $0.0044\pm 0.0021$& $1.32\pm 1.392$& $0.58\pm 0.7236$& $0.0\pm 0.0$& $0.0\pm 0.0$& $0.0005\pm 0.0006$\\

    &  \namemodel &  $0.0463\pm 0.0351$& $8.8776\pm 9.5589$& $0.0\pm 0.0$& $9.1292\pm 10.7906$& $0.0053\pm 0.0047$& $1.8163\pm 2.4717$& $4.3265\pm 3.0864$& $0.0\pm 0.0$& $0.0001\pm 0.0003$& $0.0019\pm 0.0024$\\

\cmidrule(lr){1-12}
    
    \multirow{3}{1.2cm}{ UC Irvine} & \textit{Mean} & $1.7811\pm0.6322$& $760.3568\pm 1729.6755$& $4.7351\pm11.8932$& $5.2686\pm3.0094$& $0.9488\pm0.0315$& $83.6973\pm116.533$& $15.6486\pm9.0775$& $0.004\pm0.0097$& $0.0076\pm0.0072$& $0.0959\pm0.0657$\\

    &  \taggen & $0.1883\pm0.0867$& $33.0269\pm39.9648$& $0.8495\pm2.3461$& $1.1061\pm1.3764$& $0.0067\pm0.0048$& $15.1398\pm18.8141$& $1.4247\pm2.1941$& $0.0013\pm0.0043$& $0.0018\pm0.0024$& $0.0089\pm0.0077$ \\ 
    
     & \namemodel & $0.0999\pm0.0842$& $269.7351\pm928.8703$& $1.9027\pm5.1613$& $1.2566\pm2.3706$& $0.0144\pm0.0235$& $7.9135\pm10.1291$& $3.5243\pm3.8556$& $0.0093\pm0.0335$& $0.0037\pm0.0045$& $0.0206\pm0.0282$\\ 
     
\cmidrule(lr){1-12}

    \multirow{3}{1.2cm}{Bitcoin} &  \textit{Mean}& $1.8711\pm 0.3182$& $315.7381\pm 562.1573$& $4.6369\pm 11.539$& $3.758\pm 0.9435$& $0.9394\pm 0.0171$& $62.3333\pm 58.457$& $12.3095\pm 6.5637$& $0.0286\pm 0.0586$& $0.0193\pm 0.0188$& $0.1159\pm 0.0595$\\
    
     &  \taggen &  $0.1987\pm0.0756$& $36.5536\pm27.7456$& $0.6369\pm1.4118$& $0.469\pm0.3097$& $0.0087\pm0.0071$& $12.4405\pm11.355$& $0.7024\pm0.9671$& $0.0115\pm0.034$& $0.0039\pm0.0052$& $0.0128\pm0.0096$ \\ 
     
     &  \namemodel & $0.1319\pm 0.1033$& $49.5595\pm 69.2786$& $2.7738\pm 7.6405$& $0.5066\pm 0.5994$& $0.0106\pm 0.0095$& $8.0893\pm 7.7535$& $3.506\pm 2.7753$& $0.0226\pm 0.0461$& $0.0143\pm 0.0274$& $0.0318\pm 0.029$\\ 

\cmidrule(lr){1-12}

      \multirow{2}{1.2cm}{ Wiki} &  \textit{Mean} &$1.1591\pm 0.0569$& $45.9382\pm 47.4193$& $0.0\pm 0.0$& $13.095\pm 4.5543$& $0.9844\pm 0.0101$& $8.7661\pm 5.4673$& $63.0538\pm 12.9256$& $0.0\pm 0.0$& $0.0001\pm 0.0001$& $0.0102\pm 0.0036$\\
       &  \namemodel & $0.0779\pm 0.0495$& $27.211\pm 43.2476$& $0.0\pm 0.0$& $12.1744\pm 12.6017$& $0.0083\pm 0.0094$& $3.6075\pm 4.773$& $10.8535\pm 7.0973$& $0.0\pm 0.0$& $0.0001\pm 0.0001$& $0.0025\pm 0.0033$\\ 
    \cmidrule(lr){1-12}

       \multirow{2}{1.2cm}{ Reddit} &  \textit{Mean} &$1.6721\pm 0.0589$& $6923.6465\pm 3537.6305$& $0.0\pm 0.0$& $5.663\pm 0.3618$& $0.9157\pm 0.008$& $272.0632\pm 126.3202$& $145.5296\pm 19.5631$& $0.0\pm 0.0$& $0.0008\pm 0.0005$& $0.0223\pm 0.007$ \\   
        
         &  \namemodel &  $0.1223\pm 0.0401$& $964.0444\pm 890.4033$& $0.0067\pm 0.0817$& $1.5846\pm 0.4922$& $0.0052\pm 0.0041$& $134.2164\pm 87.4022$& $54.7796\pm 18.3995$& $0.0\pm 0.0$& $0.0007\pm 0.0005$& $0.0104\pm 0.0059$ \\
        \cmidrule(lr){1-12}
        
       \multirow{2}{1.2cm}{Ta-feng}& \textit{Mean}  &$2.9705\pm 0.5458$& $85466.1681\pm 76740.864$& $0.0\pm 0.0$& $2.7509\pm 0.4356$& $0.9277\pm 0.0073$& $4056.3277\pm 1718.632$& $73.0672\pm 20.9917$& $0.0\pm 0.0$& $0.0014\pm 0.0012$& $0.151\pm 0.0267$ \\
       
       & \namemodel& $0.2991\pm 0.168$& $49010.3109\pm 92143.8563$& $0.2353\pm 0.7414$& $0.1226\pm 0.1247$& $0.0134\pm 0.0114$& $355.4118\pm 317.0144$& $277.8655\pm 69.022$& $0.0\pm 0.0$& $0.0006\pm 0.0009$& $0.0387\pm 0.0162$\\
        
        \bottomrule
\end{tabular}
}
    \caption{{\namemodel's performance against \taggen and \dymond in terms of mean absolute error $\pm$ std. dev. across various graph statistics. For all performance metrics, lower values are better. For each statistic, we also list the \textit{Mean} value over \emph{original graph snapshots} to better contextualize the error values. We do not report the results for an algorithm if it does not complete within 24 hours. Errors smaller than five decimal places are approximated to $0$.}}
    \label{tab:result_mean}
    \vspace{-0.20in}
\end{table*}

\end{appendices}

\end{document}